\pdfoutput=1

\documentclass[11pt]{article}

\usepackage{makecell}
\usepackage{multirow} %
\usepackage{array}    %
\newcolumntype{M}[1]{>{\centering\arraybackslash}p{#1}}

\usepackage[]{acl}
\usepackage{graphicx}
\usepackage{times}
\usepackage{latexsym}
\usepackage{diagbox}
\usepackage{amsmath}
\usepackage{amssymb}
\usepackage{amsmath}
\usepackage{caption}
\usepackage{threeparttable}
\usepackage{makecell}
\usepackage{cleveref}
\DeclareCaptionFormat{myformat}{\raggedright#1#2#3\par}
\usepackage[T1]{fontenc}
\usepackage{multirow}
\usepackage{array}
\newcommand{\specialcell}[2][c]{%
  \begin{tabular}[#1]{@{}c@{}}#2\end{tabular}%
}

\usepackage{graphicx} %
\newcommand{\titleimage}{\protect\includegraphics[height=0.8em]{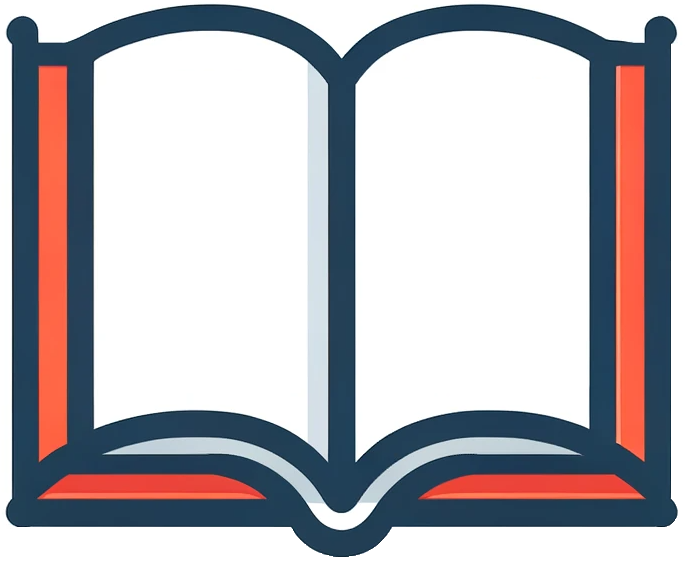}~\texttt{Book2Dial}}

\usepackage[utf8]{inputenc}

\usepackage{microtype}

\usepackage{inconsolata}

\title{\titleimage: Generating Teacher-Student Interactions from Textbooks for Cost-Effective Development of Educational Chatbots
}

\author{
    Junling Wang$^{1}$ \quad
    Jakub Macina$^{1, 2}$ \quad
    Nico Daheim$^{3}$ \\
    \textbf{
    Sankalan Pal Chowdhury$^{1}$ \quad 
    Mrinmaya Sachan$^{1}$
    } \\ \text{} \\
  $^{1}$Department of Computer Science, ETH Zurich \quad \\
  $^2$ ETH AI Center \\
  $^{3}$ Ubiquitous Knowledge Processing Lab (UKP Lab), Department of Computer Science\\ and Hessian Center for AI (hessian.AI), TU Darmstadt \\
}

\begin{document}
\maketitle
\begin{abstract}
Educational chatbots are a promising tool for assisting student learning. However, the development of effective chatbots in education has been challenging, as high-quality data is seldom available in this domain. In this paper, we propose a framework for generating synthetic teacher-student interactions grounded in a set of textbooks. Our approaches capture one aspect of learning interactions where curious students with partial knowledge interactively ask a teacher questions about the material in the textbook. We highlight various quality criteria that such dialogues should fulfill and compare several approaches relying on either prompting or fine-tuning large language models. We use synthetic dialogues to train educational chatbots and show benefits of further fine-tuning in different educational domains. However, human evaluation shows that our best data synthesis method still suffers from hallucinations and tends to reiterate information from previous conversations. Our findings offer insights for future efforts in synthesizing conversational data that strikes a balance between size and quality. We will open-source our data and code.

\hspace{.5em}\includegraphics[width=1.25em,height=1.25em]{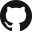}\hspace{.75em}\parbox{\dimexpr\linewidth-2\fboxsep-2\fboxrule}{\url{https://github.com/eth-lre/book2dial}}

\end{abstract}

\section{Introduction}
Educational chatbots offer a scalable way to improve learning outcomes among students \citep{kuhail2023interacting}. However, building educational chatbots has been challenging as high-quality data involving teachers and students is difficult to obtain due to various practical reasons such as privacy concerns \citep{macina-etal-2023-mathdial}. In response to this, we study the task of generating synthetic teacher-student interactions from textbooks. We create a novel dataset of textbooks drawn from an open publisher of student textbooks and present a framework (\titleimage) to generate synthetic teacher-student interactions from these textbooks. 

Our teacher-student interactions take the form of conversational question-answering (QA) interactions \citep{choi-etal-2018-quac,reddy2019coqa} where curious students ask teachers questions about the textbook and teachers answer these questions based on the textbook. Our approach primarily extends the direct instruction phase of teaching factual knowledge by facilitating more interactive information exchanges with student \cite{chi2014icap}. This is different from techniques with a focus on teaching using scaffolding \citep{macina-etal-2023-mathdial} or Socratic questioning \citep{shridhar-etal-2022-automatic} which leads to deeper and wider human learning. However, the task of generating high-quality synthetic dialogue data is difficult \citep{dai2022dialog}. This is amplified in education, where the generated interactions should cover the teaching material in an informative and coherent way. Therefore, it is important to have quality controls on such data, because students might otherwise receive wrong feedback, which could be detrimental to learning.

\begin{figure*}[]
    \centering
    \includegraphics[width=\textwidth]{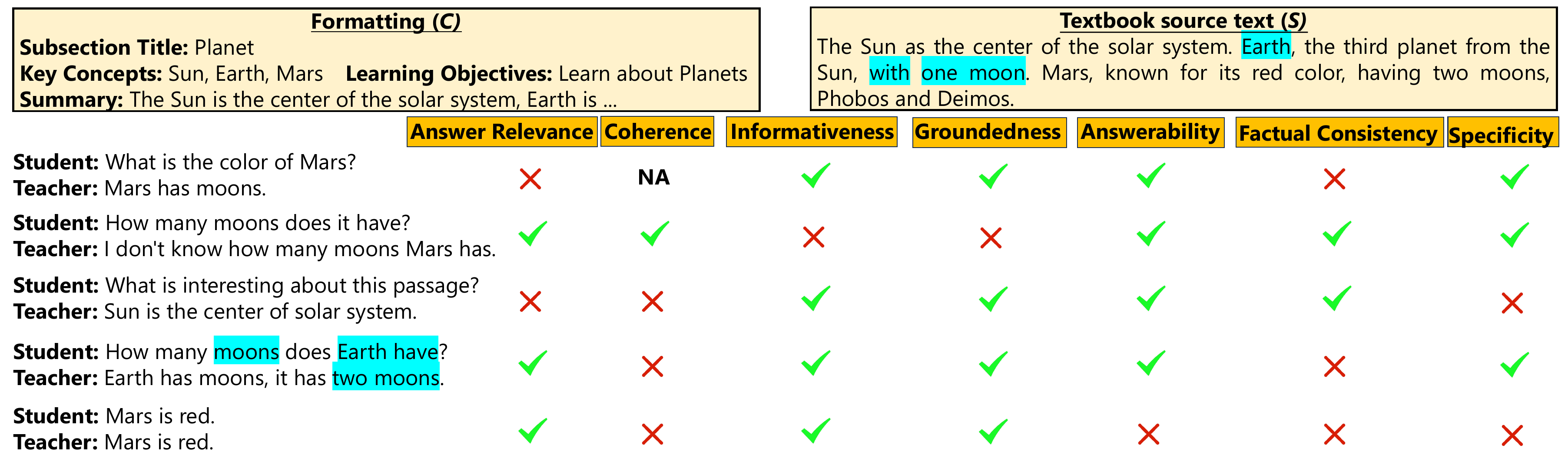}
    \vspace{-20pt}
    \caption{\small{Example of a synthetic teacher-student interaction based on a textbook, along with various criteria for evaluating the quality of the interaction. The criteria include \textit{Answer Relevance} of the answer to the question, \textit{Coherence} of the question-answer interaction with respect to the dialogue history, \textit{Informativeness} of the overall interaction, \textit{Groundedness} to the textbook, \textit{Answerability} of the question from the textbook, \textit{Factual Consistency} of the answer with respect to the question, and \textit{Specificity} of the question. More details %
    in Section \ref{sec:criteria}.}
    }
    \label{fig:fig1}
    \vspace{-15pt}
\end{figure*}

Thus, in this work, we also formulate some of the criteria that measure the quality of educational dialogues.
For example, it is crucial that the chatbot does not provide students with incorrect information and stays grounded in the textbook, ensuring factual consistency with the knowledge taught. This is particularly important given that large language models (LLMs) are prone to `hallucinations' or generating plausible but incorrect or unverified information \citep{ji2023survey}. 
While a simple teacher strategy would be to just answer with extracted passages from the textbook, this might hurt the coherence of the dialogue which is present in interactive educational situations \citep{baker2021educational}. The teacher's response should both be relevant to the student's question \citep{ginzburg2010relevance}, as well as, informative as this ensures that key information from the textbook is covered in the dialogue \citep{tan2023does}. We formalize these requirements into criteria shown in Figure \ref{fig:fig1}. 

\titleimage\ comprises of three approaches taken from prior work: 
multi-turn QG-QA \citep{kim2022generating}, Dialogue Inpainting \citep{dai2022dialog} and prompting LLMs to role-play teacher and student. We use the formatting information in the textbook, such as titles or key concepts, to initialize student models with imperfect information. In contrast, the teacher models have perfect information and are expected to generate grounded responses based on the textbook. We fine-tune and prompt various open-source language models to generate teacher-student interactions.

We evaluate \titleimage\ on the proposed quality criteria and use human evaluations to support our findings. Results reveal that data generated by role-playing LLMs scores highest in most criteria, as shown in \Cref{sec:stata_metrics} and \ref{sec:metric_result}, demonstrating reasonable efficacy in creating educational dialogues. The generated dialogues effectively incorporate textbook content but fall short in mimicking the natural scaffolding of educational conversations and exhibit issues like hallucinations and repetition, as discussed in \Cref{sec:qualitative}. Despite these limitations, we show that the generated synthetic data may be used to pre-train educational chatbots with benefits in some educational domains (\Cref{sec:downstreamtask}), especially when the domain of the evaluation dataset matches with the pretraining data.

\begin{figure*}[h!]
    \centering
    \includegraphics[width=0.99\textwidth]{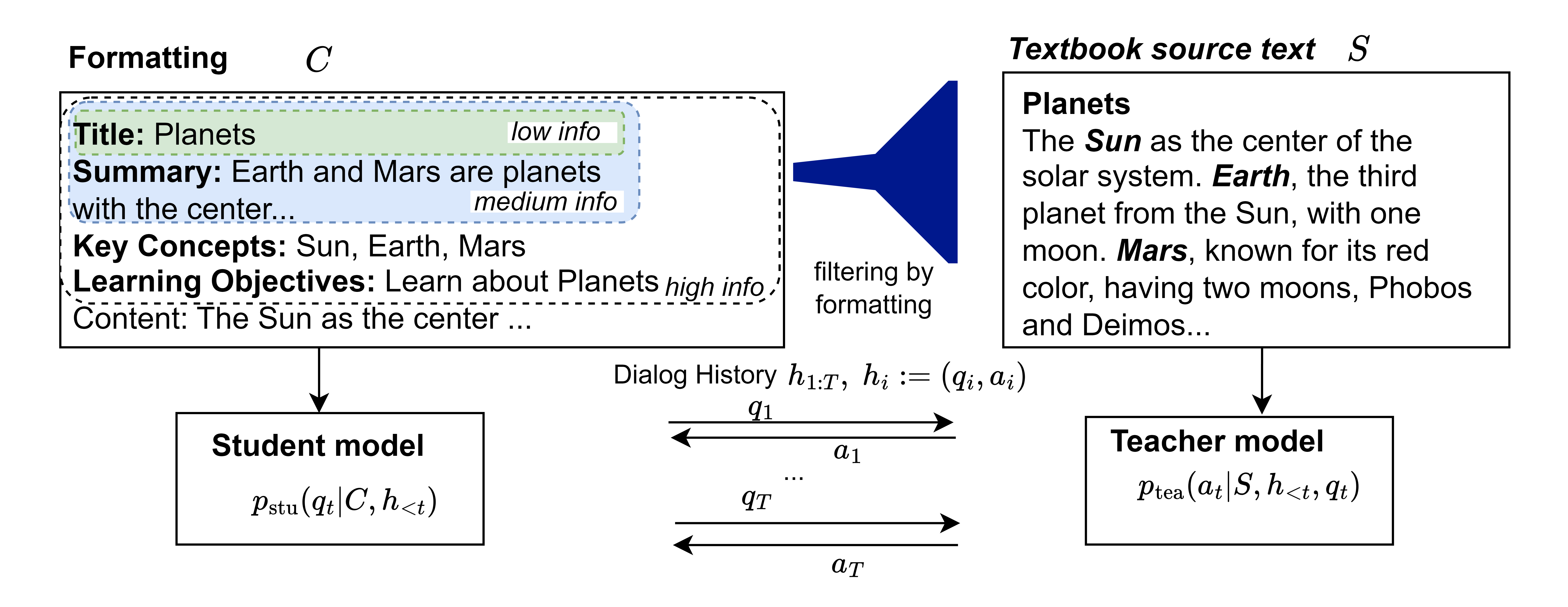}
    \vspace{-10pt}
    \caption{\small{\titleimage\ Framework for Generating Dialogues from Textbooks: Our approach uses two models -- a Student model and a Teacher model. The Student model plays the role of a student, formulating questions from a limited context (document formatting). In contrast, the Teacher model assumes the role of a teacher, providing answers and guidance by referencing the (sub-)section in the textbook. This framework can be adapted to various instantiations of the two roles with varying formatting information, such as multi-turn QA-QG models \citep{kim2022generating}, Dialogue Inpainting \citep{dai2022dialog}, and a prompting approach utilizing role-playing LLMs.
    } }
    \label{fig:fig2}
    \vspace{-10pt}
\end{figure*}

\section{Related Work}

\subsection{Synthetic Data for Conversational QA}
Prior work in educational research has focused on generating individual questions \citep{kurdi2020systematic} under two common settings: answer-aware and answer-unaware generation. The former approach starts by identifying an answer and then generates a question accordingly, whereas the latter generates a question without pre-determining the answer. These approaches have also been extended to generating multiple questions \citep{rathod2022educational}, causal question generation \citep{stasaski2021automatically}, prediction of question types to ask \citep{do-etal-2023-modeling}, or decomposing problems into Socratic subquestions \citep{shridhar-etal-2022-automatic}. However, most works do not address conversational settings.

Datasets like QuAC \citep{choi-etal-2018-quac} and CoQA \citep{reddy2019coqa} focus on conversational question answering in non-educational settings. Previous work has also explored strategies for creating such data with humans or automatically by using models.
For example, \citet{qi-etal-2020-stay} withholds the context required for answers from the questioner, leading to information-seeking questions. SimSeek \citep{kim2022generating} synthesizes datasets for conversational question answering from unlabeled documents but do not demonstrate significantly improved performance in downstream tasks. A recent work, Dialogizer \citep{hwang-etal-2023-dialogizer}, proposes a framework for generating context-aware conversational QA dialogues. However, these methods do not take into account the specifics of the educational domain.

Other works outside of conversational question answering also explore data synthesis. For example, \citet{kim2022soda} source dialogue data from a commonsense knowledge graph, \citet{chen2023places} prompt LLMs, and \citet{bao-etal-2023-synthetic} create grounded dialogue datasets.

\subsection{Educational Dialogue Datasets}
The development of educational chatbots is highly reliant on quality data. Yet such data is hard to obtain. %
Therefore, previous works such as MathDial \citep{macina-etal-2023-mathdial} collect conversational data by pairing real teachers with an LLM that simulates students.
Other datasets are commonly created by roleplaying both teacher and student, such as CIMA \citep{stasaski2020cima} or by transcribing classroom interactions \citep{suresh-etal-2022-fine, demszky-hill-2023-ncte} or logging online conversations \citep{caines2020teacher}.
However, all of these methods are challenging to scale, and using non-experts often leads to data quality issues \citep{macina-etal-2023-mathdial}.

Thus, in this work, we explore data synthesis as a scalable way of creating such data. Data augmentation and synthetic data generation have gained attention as effective techniques to overcome the challenges associated with manual data annotation. Synthetic data generation has been shown to be a promising approach. For instance, \citet{kim2022soda} demonstrated the potential of sourcing dialogue data from common sense knowledge. However, ensuring the objectivity of generated data remains a concern. Similarly, \citet{zhang2018synthetic} introduced innovative methods for task-oriented dialogue synthesis. However, its dependency on predefined schemas limits its scalability.

\section{Educational Conversation Generation}
We first introduce a framework for dialogue synthesis from textbooks in Section \ref{sec:task}, and then discuss the quality criteria that the generated dialogues should fulfill in Section \ref{sec:criteria}.

\subsection{\titleimage\ Framework}\label{sec:task}

We set out to create meaningful teacher-student interactions from educational textbooks in the form of conversational QA pairs between the teacher and the student. In order to generate these interactions, we assume that the ``teacher'' is knowledgeable about the textbook content, and the ``student'' only knows limited information from the textbook. Thus, we intuitively provide the {\bf teacher model} all the textbook information but withhold some information from a {\bf student model}.
For this, we can use the structuring and formatting elements found in textbooks, including 1) \textbf{Titles:} headings of sections and subsections; 2) \textbf{Summary:} summaries of chapters; 3) \textbf{Other Metadata:} key concepts, learning objectives, bold terms, and the introductory paragraph of each section; and assume that the student model only has access to this information.

During the conversation, the ``student'' asks inquisitive questions about the textbook while the ``teacher'' guides them by answering these questions and including additional information in their response. 
Formally, a dialogue $d$ comprises of a sequence of $T$ question-answer interactions: $d = \{(q_1, a_1), \dots, (q_T, a_T)\}$.
The formalization of the task is depicted in Figure \ref{fig:fig2}. The student model $p_{\text{stu}}(q_t|C, h_{<t})$ generates a question $q_t$ given the dialogue history $h_{<t} = \{(q_i,a_i)\}_{i=1}^{t-1}$ and the partial context (formatting information) $C$. The teacher model $p_{\text{tea}}(a_t|S, h_{<t}, q_t)$ generates the answer response $a_t$ given the question, the dialogue history and the full textbook source $S$.

\subsection{Evaluation of Educational Conversations}\label{sec:criteria}

To build a high-quality educational conversation, we focus on the student to ask questions that are \textbf{specific} enough to drive the conversation forward, and also \textbf{answerable} given the context. The teacher must then respond with an answer that is \textbf{relevant} to the question, \textbf{factually consistent} with the context, and \textbf{informative} to the student. Finally, the overall conversation should be \textbf{coherent} and \textbf{grounded} to the entire context, not just parts of it. We use this as our guiding principle and define $7$ criteria to evaluate the quality of a good educational interaction. 
Although these metrics may not necessarily be mutually exclusive or extensively address all possible features of high-quality educational dialog, each of them serves as an important aspect to be considered in the education domain. We selected those that could be instantiated by measures and are potentially useful to supplement the traditional factual classroom listening to a lecture with interactive chatbot interactions \citep{chi2014icap,ruan2019quizbot}. We detail these criteria in the rest of this subsection.

\subsubsection{Answer Relevance}
Answer Relevance measures how directly related the answer is to the question in each QA pair in the dialogue. This criterion is crucial in education, as teachers should have subject-matter knowledge and can adapt responses to the student's various knowledge levels \citep{lepper2002wisdom}. 

We assess the Answer Relevance of individual QA pairs and then combine these assessments to determine the dialogue's overall Answer Relevance. We use {\bf BF1($q_t$, $a_t$)}, {\bf QuestEval} and {\bf Uptake} as metrics for Answer Relevance. The BF1 metric uses BERTScore F1 \citep{bert-score}  for semantic alignment between questions and answers using BERT embeddings. QuestEval \citep{scialom-etal-2021-questeval} generates questions from both the question and answer, then generates answers for these questions, comparing them to measure relevance. The Uptake metric \citep{demszky-etal-2021-measuring}, specific to the education domain, analyzes teachers' responses to student utterances, focusing on their dependence and relevance. More details are in Table \ref{tab:criterion_formula}.

\subsubsection{Coherence of the Dialogue}\label{sec:coherence}
Coherence measures whether QA pairs in the dialogue form a logical and smooth whole, rather than independent QA pairs. Coherence is an important aspect of good dialogue \citep{dziri2019evaluating} and it is important in education to sequentially instruct students with the new textbook knowledge in a coherent way.

We adapt two metrics, {\bf BF1($q_t$, $a_{<t}$)} and {\bf BF1($q_t$, $a_{(t-1)}$)}, to measure coherence, similar to the approach in \citep{kim2022generating}. The first metric uses BERTScore F1 to evaluate the current question against each previous answer as references, whereas the second metric compares the current question solely with the immediately preceding answer using BERTScore F1. See Table \ref{tab:criterion_formula} for more details.

\subsubsection{Informativeness}
Informativeness evaluates the amount of new information introduced by each student-teacher interaction in the dialogue. This criterion is important to analyze the rate at which the textbook content is progressively introduced to the student.

To assess Informativeness, we use \textbf{1 - Overlap($a_t$, $a_{<t}$)}, calculating one minus the ratio of token intersection over union in the current and all previous answers for each QA pair. 
This metric is correlated with human evaluation, as detailed in Appendix \ref{sec:metric_eval} and Table \ref{tab:criterion_formula}.

\subsubsection{Groundedness to the Textbook}
This criterion assesses the amount of information from the textbook incorporated into the dialogue. 
Two metrics are used for assessment: {\bf Density}, evaluating the average length of text spans extracted from textbook content $S$ and included in the dialogues; {\bf Coverage}, measuring the proportion of dialogue words originating from the textbook. Both metrics are adopted from \citep{grusky-etal-2018-newsroom}, and their formulas are shown in Table \ref{tab:criterion_formula}.

\subsubsection{Answerability of the Questions}
Answerability measures whether the student's question is answerable given the textbook content. While non-answerable questions could also represent the curiosity of a student \cite{scialom-staiano-2020-ask}, the answerability of questions given the context is generally important for a more useful dialog.
We use the ``distilbert-base-cased-distilled-squad'' QA model\footnote{https://huggingface.co/distilbert-base-cased-distilled-squad} to judge whether each question is answerable given the textbook content, and refer to this metric as \textbf{Answerability}. This approach is akin to the method employed in \citep{kim2022generating}. More details are in Table \ref{tab:criterion_formula}.

\subsubsection{Factual Consistency of the Answer}
Factual Consistency measures whether the answer correctly responds to the student's question. This criterion is crucial in education because it is important for students to learn accurate information \citep{metzger2003college}.
Existing metrics like $Q^2$ \citep{honovich2021q} use a QA model to assess answer correctness, while RQUGE \citep{mohammadshahi2022rquge} uses a QA model to evaluate the quality of the candidate question. In our scenario, we need to measure whether the answer contains correct information and accurately answers the question. Therefore, we build on the idea of $Q^2$ and introduce a new metric referred as {\bf QFactScore}: 
\begin{equation}
\alpha\cdot\text{sim}(\text{QA}(q_t, S) , a_t)  + \beta\cdot\text{sim}(q_t, a_t)
\end{equation}

It calculates the cosine similarity of embeddings between the predicted and original answers for each QA pair and also evaluates the similarity between the question and the original answer. This metric has been validated for its alignment with human evaluation, as detailed in Appendix \ref{sec:metric_eval}. The final score is the weighted sum of two similarity scores. More details are in Table \ref{tab:criterion_formula} and Appendix \ref{sec:qfactscore}.

\subsubsection{Specificity of the Question}
Specificity assesses whether the question is specific, rather than general. An example of a generic question is `What is interesting about this passage?'. 
We assess specificity through human evaluation, as there is no existing metric that captures specificity.

\section{From Textbooks to Dialogues}\label{dialog_gen}
In this section, we describe different methods used for generating dialogues from educational textbooks in \titleimage, namely:

\begin{enumerate}
\itemsep0em 
    \item {\bf Multi-turn QG-QA models.} In this setting, we use fine-tuned QG and QA models interacting with each other. %
    \item {\bf Dialogue Inpainting.} \citet{dai-etal-2021-dialogue} uses a span extraction model over the textbook as a teacher model, where the response is copied from the textbook and the question is generated by a QG model acting as the student.
    \item {\bf Persona-based Generation.} This approach uses LLMs like GPT-3.5, and leverages prompting to interactively simulate the student and the teacher and generate dialogues.
\end{enumerate}

We describe the implementation of these methods below. More details are in Appendix \ref{sec:textbook_to_dialogues_detail}.

\subsection{Multi-turn QG-QA models}

This scenario utilizes separate QG and QA models to interact in a multi-turn scenario. As a representation of this approach from related work, we consider the SimSeek-asym model \citep{kim2022generating}. 
The approach consists of two components: 
\begin{enumerate}
\itemsep0em 
    \item A {\bf Question Generation} (QG) model for generating conversational questions relying solely on prior information (i.e., formatting information relevant to the topic). This student model generates question based on the dialogue history and filtered Information $C$: $p(q_t | C, h_{<t})$.
    \item A {\bf Conversational Answer Finder} (CAF) to comprehend the generated question and provide an acceptable teacher answer to the question from the evidence passage: $p(a_t | S, h_{<t},q_t)$.
\end{enumerate}

\subsection{Dialogue Inpainting}
Dialogue Inpainting \cite{dai2022dialog} is an approach for dialogue generation characterized by its information-symmetric setting.
In this framework, both the student and teacher model are provided with the complete textbook text $S$. The teacher model simply iterates over each sentence in $S$ and copies it as an answer. The student model is a QG model. We use data from the OR-QuAC \citep{qu2020open}, QReCC \citep{anantha-etal-2021-open}, and Taskmaster-2 \citep{taskmaster2_2020} datasets to train the student model. For the student model, a dialogue reconstruction task is employed. At training time rather than distinguishing questions and answers, the dialogue reconstruction task treats a conversation as a sequence of utterances $\{u_i\}_{i=1}^{2T}$,  To train it, a randomly chosen utterance $u_i$ is masked to create a partial dialogue $d_{m(i)}$ = $u_1, \cdots, u_{i-1}$, <mask> , $u_{i+1}, \cdots, u_{2T}$. The model then predicts $u_i$ and is trained by minimizing the loss:

\begin{equation}\label{eq}
    \mathcal{L}(\theta) = - \sum_{d \in D} \mathbb{E}_{u_{i} \sim d}[ \log p_{\theta}(u_{i} \mid d_{m(i)})]
\end{equation}

\noindent During inference, the model uses each sentence in the textbook as a teacher's utterance and only predicts student utterances accordingly, $\{u_{2k-1}\}_{k=1}^{T}$ corresponding to $\{q_i\}_{i=1}^{T}$ in our notation. We basing our model (eq \ref{eq}) on FLAN-T5-XL \citep{chung2022scaling}. More details are in Appendix \ref{sec:dialogue_inpainting_detail}.

\subsection{Persona-based Generation}\label{sec:llm-based_generation}

Inspired by \citep{markel2023gpteach}'s idea of using LLMs to simulate student personas, we propose a method to simulate student and teacher personas by prompting LLMs for dialogue generation.
We use one instance of GPT-3.5 model to play the student and another to play the teacher.\footnote{We used the GPT-3.5-turbo API between 25th September and 4th October, 2023.} The teacher model is provided with all the information from the textbook, including content and all the formatting information. The information provided to the student model is varied. We consider four variants for generating dialogue in each subsection listed with an increasing amount of information provided to the student model: 1) \textit{Persona (Low Info)} provides the student model with only the Title information, 2) \textit{Persona (Medium Info)} provides both the Title and Summary information, 3) \textit{Persona (High Info)} offers all formatting information, and 4) 
\textit{Persona (Single Instance)} generates the entire dialogue using a single LLM instance, equipping one model with formatting and textbook content information. More details are in Table \ref{tab:model_comparison}.

Considering that GPT-3.5 is proprietary and not open-source, we adopted prompting techniques to steer the models in dialogue generation. The prompt for Persona (High Info) and Persona (Single Instance) is detailed in Appendix \ref{sec:prompt}.

\section{Results and Analyses}
In this section, we aim to address the following research questions:

\begin{enumerate}
\itemsep0em 
    \item How does the choice of generation framework influence the quality of the generated data?
    \item What is the optimal amount of information that should be incorporated into the student model to produce natural dialogues?
    \item How does pre-training on the synthesized data in various domains improve the performance of models that are fine-tuned and evaluated on existing datasets?
\end{enumerate}

To address these questions, we generate dialogues from textbooks across various domains and analyze the generated dataset. 

\noindent{\bf Textbook data:} We collected 35 textbooks available on OpenStax\footnote{https://openstax.org/}, spanning domains of math, business, science, and social science. From these, we select four textbooks to create our dialogue datasets. 
Table \ref{tab:textbook_statistics_4} provides statistics of the four textbooks. The first and second research questions are addressed in Sections \ref{sec:model_results_comparison} and \ref{sec:human_eval}, respectively, while the third question is answered in Section \ref{sec:downstreamtask}. 

\subsection{Automatic Evaluation}\label{sec:model_results_comparison}
In this section, we discuss statistics and metrics for the generated datasets. We present aggregated statistics of the datasets across four textbook domains in Tables \ref{tab:metric_table1} and \ref{tab:metric_table2_new}, also noting comparisons with existing datasets: MathDial \citep{macina-etal-2023-mathdial} -- math tutoring conversations between human teachers and student LLMs, fact-based human-human conversations of QuAC \citep{choi-etal-2018-quac}, and NCTE transcripts of math classrooms \citep{demszky-hill-2023-ncte}. Domain-specific results in the dataset are detailed in Tables \ref{tab:statistic_table_full} and \ref{tab:metric_table_full}. To adjust for varying dialogue lengths, we limited the number of turns to $T = 12$ for each model, as in \citep{kim2022generating}.

\subsubsection{Statistical Analysis of the Datasets}\label{sec:stata_metrics}

In dialogue, different types of questions emphasize various aspects. We hypothesize that "what" and "which" questions focus on factual information. In contrast, other question types, such as "why" and "how," tend to reflect more complex inquiries, which are also important in educational contexts. In Table \ref{tab:metric_table1}, we present the percentages of student questions including words \textit{what}, \textit{which}, \textit{why}, and \textit{how}\footnote{This ratio excludes 'how much' and 'how many' questions because they pertain to factual information.}. Furthermore, the average token count for questions and answers across each dataset is also shown. 
The key findings are as follows:

\paragraph{Less factual questions in the Persona (Single Instance) dataset} The Persona (Single Instance) model generates the fewest "what" or "which" questions compared to other synthesized datasets, suggesting more diverse questioning. It also has a similar question type distribution to MathDial.

\paragraph{More ``how'' questions in Persona datasets} The four Persona datasets contain the highest ratio of `how' questions, which suggests a higher ratio of questions asking for explanations.

\paragraph{High token counts in Persona datasets} Datasets from Persona models feature the high average token counts in questions and answers, suggesting these dialogues are more verbose and informative. NCTE features high token counts in questions, typical in classroom transcripts with lengthy teacher inquiries and brief student responses.

\begin{table}
\centering
\resizebox{\linewidth}{!}{
\begin{tabular}{l|ccc|cc}
 & \multicolumn{3}{c|}{\textbf{Question Type (\%)}} & \multicolumn{2}{c}{\textbf{Num. Tokens in:}}\\
\cline{2-6}
& \textbf{ \specialcell{ what /\\which}} & \textbf{ \specialcell{ why}} & \textbf{ \specialcell{ how}}  & \textbf{ \specialcell{Questions}} & \textbf{ \specialcell{Answers}}\\
\hline

SimSeek  & 55.00 & 2.00 & 17.50  & 10.90 & 14.33 \\
\hline
Dialogue Inpainting & 63.50  & 2.75 & 16.00  & 6.85 & 19.63\\
\hline
Persona (Single Inst.) & 28.25 & \textbf{5.50} & 23.25  & 14.97 & 34.58 \\
Persona (Low Info) & 70.25  & 0.75 & \textbf{27.50}  & 17.56 & 84.75\\
Persona (Med. Info) & 69.00 & 0.25 & 27.00  & 17.69 & 85.19\\
Persona (High Info) & \textbf{73.50} & 0.75 & 24.00  & 19.01 & 84.70 \\
\hline\hline
MathDial & 21.00 & 5.00 & 10.00 & 17.11 & 32.91 \\
QuAC & 36.00 & 3.00 & 8.00 & 6.52  & 12.62\\
NCTE & 40.00 & 4.00 & 10.00 & 33.85 & 4.41
\end{tabular}
}
\caption{
Key statistics of the synthesized educational dialogue dataset: Persona-generated datasets contain more 'what/which' and 'how' questions, and the dialogues tend to be more verbose.
}
\label{tab:metric_table1}
\end{table}

\begin{table*}
\centering
\scriptsize
\begin{tabular}{p{2cm}|M{0.6cm}M{0.8cm}M{0.6cm}|M{1.5cm}|M{0.6cm}M{0.8cm}|M{0.8cm}M{1cm}|M{1.3cm}|M{1.1cm}}
 & \multicolumn{3}{c|}{\textbf{\specialcell{Answer\\Relevance}}} & \textbf{\specialcell{Informativeness}} & \multicolumn{2}{c|}{\textbf{\specialcell{Groundedness}}} & \multicolumn{2}{c|}{\textbf{\specialcell{Coherence}}} &\textbf{\specialcell{Answerability}} & \textbf{\specialcell{Factual\\Consistency}} \\
\cline{2-11}
& \textbf{\specialcell{BF1\\($q_t$, $a_t$)} } & \textbf{QuestEval } & \textbf{Uptake} & \textbf{\specialcell{1 - Overlap\\($a_t$, $a_{<t}$)} }  & \textbf{Density } & \textbf{Coverage } & \textbf{\specialcell{BF1\\($q_t$,  $a_{<t}$)} } & \textbf{\specialcell{BF1\\($q_t$, $a_{t-1}$)} }  & \textbf{Answerable } & \textbf{QFactScore } \\
\hline
 SimSeek & 0.53 & 0.25 & 0.78 & 0.71  & 11.66 & 0.82 & 0.51 & 0.55  & 0.84 & 0.32 \\
\hline
 Dialogue Inpainting & 0.52 & 0.28 & 0.84 & \textbf{0.91}  & \textbf{22.62} & \textbf{0.90} & 0.45 & 0.46  & 0.75 & 0.24 \\
\hline
 Persona (Sing. Inst.) & 0.58 & 0.35 & 0.98 & 0.86 &  3.94 & 0.75 & 0.49 & 0.52   & 0.92 & 0.54 \\
 Persona (Low Info) & 0.61 & \textbf{0.44} & \textbf{0.99} & 0.59  & 2.39 & 0.70 & 0.52 & \textbf{0.59}  &0.98 & 0.75 \\
 Persona (Med. Info) & 0.61 & \textbf{0.44} & \textbf{0.99} & 0.59  & 2.43 &  0.71 & 0.52 & \textbf{0.59} & \textbf{0.99} & {\bf 0.76} \\
 Persona (High Info) & \textbf{0.62} & \textbf{0.44} & \textbf{0.99} & 0.60 & 2.50 & 0.71 & \textbf{0.53} & \textbf{0.59} & \textbf{0.99} & 0.75 \\
 \hline\hline
 MathDial & 0.46 & 0.30 & 0.83 & 0.64 & 1.30 & 0.46 & 	0.42 & 0.47 & 0.51 & 	0.39 \\
 QuAC &  0.43 & 0.24 & 0.76 & 0.72 & 13.78 & 0.81  & 0.42 & 0.43 & 0.73 & 0.38 \\
 NCTE & 0.34 & 0.21 & 0.76 & 0.89 & NA & NA & 0.38 & 0.37 & NA & NA \\
\end{tabular}

\caption{
Automatic evaluation metrics for the synthesized dialogue data and existing datasets. Higher values indicate better performance. Persona-generated dialogues score highest in Answer Relevance, Coherence, Answerability, and Factual Consistency, while Dialogue Inpainting generated dialogues score highly in Informativeness and Groundedness.
}
\label{tab:metric_table2_new}
\end{table*}

\subsubsection{Data Quality Metrics}\label{sec:metric_result}
We report the various data quality metrics in Table \ref{tab:metric_table2_new}. Our key findings are as follows:

\paragraph{Persona datasets are solid in most of the criteria} 
Persona models generated datasets outperform others in most metrics, indicating their good ability in creating dialogues from textbooks. Note the MathDial, QuAC, and NCTE datasets each focus on different domains with different data collection designs and grounding texts, therefore they are not directly comparable in these metrics and should be interpreted with their specific context in mind. For example, a tutoring dataset like MathDial is expected to have lower answerability as tutors tend to ask challenging questions while the specific data collection of QuAC to include supporting arguments increases the density.

\paragraph{High Informativeness and Groundedness of Dialogue Inpainting dataset:} Dialogue Inpainting models achieve the highest score across all models in Informativeness and Groundedness. This is expected as this model uses sentences in the textbooks as teachers' answers.

\paragraph{Students with more information access perform better in automatic metrics.} 
Datasets from Persona (High Info) and Persona (Medium Info) typically outperform or match with datasets from Persona (Low Info). This suggests that more information to the student may enhance key criteria.
However, the impact differences among formatting levels are not markedly significant, indicating a need for further research on this question.

\subsection{Human Evaluation}\label{sec:human_eval}
To compensate for the limitations of automatic metrics, human evaluations of SimSeek, Dialog Inpainting, and Persona (High Info) dialogues were conducted based on seven criteria: Answer Relevance (AnsRel), Informativeness (Info), Groundedness (Gro), Coherence (Coh), Factual Consistency (Fact), Answerability (Ans), and Specificity (Spe). Questions for judging each criterion are in Table \ref{tab:human_eval_criterion}.
We recruited 4 annotators to evaluate 12 dialogues each, yielding an average Cohen's Kappa of 0.74, indicating reasonable agreement. Evaluation details are in Appendix \ref{sec:human_eval_detail}, and results in Table \ref{tab:human_eval_result1}.

\begin{table}[h!]
\centering
\resizebox{\columnwidth}{!}{
\begin{tabular}
{l|ccccccc}
& \textbf{AnsRel} & \textbf{Info} & \textbf{Gro} & \textbf{Coh}  & \textbf{Fact} & \textbf{Ans} & \textbf{Spe} \\
\hline
\specialcell{SimSeek} & 0.32 & 0.56 & \textbf{1.00} & 0.58 & 0.25 & 0.66 & 0.89\\
\specialcell{Dial. Inpaint.}  & 0.58 & \textbf{0.97} & \textbf{1.00}  & 0.66 & 0.73 & 0.83 & 0.61\\
\specialcell{Persona\\(High Info)} & \textbf{0.97} & 0.74 & 0.99  & \textbf{0.85} & \textbf{0.79} & \textbf{0.96} & \textbf{0.99} %
\end{tabular}
}
\caption{Human Evaluation Result: Persona (High Info) generated dialogues score highest in Answer Relevance, Coherence, Factual Consistency, Answerability and Specificity, while Dialogue Inpainting generated dialogues excel in Informativeness and Groundedness.}
\label{tab:human_eval_result1}
\end{table}

Persona (High Info) excels among the three models, leading in Answer Relevance, Coherence, Factual Consistency, Answerability, and Specificity, rendering it the most suitable choice based on our dialogue generation objectives. This result aligns with the results of automatic metrics presented in Table \ref{tab:metric_table2_new}. 
However, the dialogues generated by the Persona-based method exhibit only an average score in Informativeness, with a score of 0.74 indicating that approximately 26\% of QA pairs do not contribute new information to the conversation. 
The Persona-based model, while leading in Factual Consistency among the three models, scores only 0.79, which indicates that approximately 21\% of the QA pairs lack Factual Consistency. For educational dialogues, it's imperative to aim for high Factual Consistency to ensure the reliability of the knowledge imparted. The primary reason for this issue is the hallucination in LLMs, where LLMs respond to questions using fabricated or false information not grounded in the textbook. This poses a significant challenge and calls for further research into ways to better ground LLMs to text documents in the future.

\subsection{Qualitative Human Analysis}\label{sec:qualitative}
We inspected dialogues generated by each model and found several limitations compared to natural educational conversations. We report qualitative descriptions of these findings:

\paragraph{Insufficient follow-up ability of Persona models}
Even though Persona models outperform other methods in automatic and human evaluations, the model has several issues. Dialogues generated by Persona models are unlike natural conversations and resemble a series of artificial QA pairs about textbooks. Moreover, the dialogue does not have enough follow-up questions by students and rather broadly touches on the textbook content rather than going into depth about certain topics.

\paragraph{Repeating answers in SimSeek and Persona}
In the SimSeek and Persona datasets, we find that teacher answers often reiterate information from previous interactions. SimSeek often generates questions related to the previous teacher answer, while Persona often provides summaries of textbook content in each answer.
 
\paragraph{Insufficient Specificity of Dialogue Inpainting}
In alignment with the results of human evaluation, we find that the Dialogue Inpainting model tends to generate ``general'' questions, such as ``What is interesting about this passage?'' These types of questions, which are not specific to the textbook content, are less desirable in educational dialogue.

\subsection{Pre-training for Educational Chatbots}\label{sec:downstreamtask}
In this section, we show the effectiveness of these imperfect but automatically created synthetic data based on any textbook domain for downstream task of pre-training educational chatbots. We pre-train simple chatbot models with synthesized data and assess their performance on educational conversation tasks.

Specifically, we use text generation models based on language models to generate teacher responses $a_t$ given the dialogue history $h_{<t}$, textbook grounding information $S$ and the question $q_t$.
We compare two scenarios: (1) a model pre-trained on synthetic datasets, then fine-tuned and tested on various educational or information-seeking dialogue datasets; and (2) a model trained and tested solely on these dialogue datasets without pre-training. We used FLAN-T5-LARGE \citep{chung2022scaling} as our base language model. For our test sets, we use the MCTest and CNN splits of the CoQA dataset \citep{reddy2019coqa}, as well as the NCTE dataset \citep{demszky-hill-2023-ncte}. The MCTest split contains dialogues about children's stories; the CNN split contains conversations about the news; the NCTE dataset contains transcripts of elementary math classrooms. 

We pre-trained the base model on four textbook-based synthetic datasets, each from a different subject: math, business, science, and social science. The datasets and training details are shown in Appendix \ref{sec:downstream_task_detail}. The results are shown in Table \ref{tab:chatbot}. We report the BLEU score of the scenario where we pre-trained the base model on our textbook-generated dialogue dataset and the difference between this pre-train version against the version without this pre-training shown in brackets.

We found that the model that was first pre-trained on the social science textbook data achieved the highest score when tested on MCTest and CNN splits of the CoQA dataset, with improvements of 4.16 and 1.99. Meanwhile, the model pre-trained on the business textbook data achieved the highest score when tested on the NCTE dataset. The model pre-trained on the math textbook data also shows improvements. As the social textbook dataset contain the least math expressions, it improves most in non-math domains but does worst in the math domain. We conclude that synthetic datasets created from textbooks may be effective for pre-training chatbots if they align with the target domain.

Upon a more qualitative human examination of the generated results, we found that the pre-trained models have a better understanding of the input context and generate more correct answers than the corresponding non-pre-trained models. Some example generations are shown in Appendix \ref{sub:Downstream_Task_Generation_Results}.

\begin{table}[h!]
\centering
\resizebox{\columnwidth}{!}{
\begin{tabular}{p{1.3cm}||c|c|c}
& CoQA (MCTest) & CoQA (CNN) & NCTE    \\
\hline\hline
Math & 26.10 (+4.03) & 13.95 (+0.82) & 8.79 (+0.39) \\ 
\hline
Business & 18.91 (-3.22) & 13.29 (+0.16) & \textbf{8.99 (+0.59)} \\
\hline
Science & 22.36 (+0.22) & 14.96 (+1.83) & 8.73 (+0.33) \\
\hline
Social & \textbf{26.30} (\textbf{+4.16}) & \textbf{15.11} (\textbf{+1.99}) &  8.37 (-0.03)\\
\hline
All &  23.05 (+0.92) & 14.31 (+1.19) & 8.41 (+0.01) %

\end{tabular}
}
\caption{
Downstream Task Results. We use dialogues generated from one textbook from each domain for pre-training and evaluate on downstream benchmarks. Each cell displays BLEU score and the (difference from the baseline), where the baseline is derived from the same model without pre-training.
}
\label{tab:chatbot}
\end{table}

\section{Conclusion}
This work introduces a new task of generating educational dialogues from textbooks to help pre-train educational chatbots in various domains where dialogue data are scarce. We detail and compare various approaches and settings to simulate student-teacher interactions and create such data. We evaluated the generated dialogues, focusing on some measures of goodness, such as Answer Relevance, Informativeness, Coherence, and Factual Consistency. Our results indicate that the approach with LLMs role-playing as teachers and students for data synthesis performs well in these metrics. However, upon closer human inspection, we also observed several issues with the synthesized data such as the problem of hallucinations and repeating information. Despite these issues, we showed that the generated dialogues could be used to pre-train educational chatbots and achieve performance improvements in various educational settings.

\section{Limitations}
\textbf{Focus on a specific teaching scenario and limitations in educational contexts} In this work, we focus on a specific educational scenario where a curious student asks questions to a knowledgeable teacher. It has been shown that the quality of the student's questions (with deep reasoning ones) is correlated with their learning \cite{graesser1994question, person1994inferring}. We did not model any of these aspects in our approach.
Furthermore, recent approaches of teachers asking Socratic questions or providing indirect scaffolds and hints instead of providing students directly with answers have also been shown to lead to better learning outcomes \citep{freeman2014active}. In our formulation, teachers directly provide students with answers. Our approach focuses on facilitating informational exchanges and is more suitable for helping students access the entire content of the textbook in a new interactive way and through their interests. This serves as a starting point for developing more sophisticated educational chatbots in the future. Future work could focus on other interaction scenarios and combine our approach with Socratic questioning \citep{shridhar-etal-2022-automatic} and scaffolding \citep{macina-etal-2023-opportunities,sonkar2023class} to achieve significantly improved applicability to educational use cases.

{\bf Achieving the highest scores in all metrics is not the overall goal for human learning.} Considering Informativeness, while a dialogue rich in information suggests a potential for a greater extent of learning by a student, there exists a trade-off, as excessive information can increase the student's cognitive load and become overwhelming \citep{mayer2003nine}. Therefore, finding the optimal amount of information that the dialogue should contain needs careful consideration in future work. Similarly, for other metrics, educational practitioners could ideally set the target metrics and their combination for achieving better pre-training performance steered towards educational use cases. 

\textbf{Aspects of evaluation framework}: Although we tried to include various aspects of the evaluation in this work, it was not feasible to focus on all important educational aspects. We specifically focused on one setting, where students ask curious questions and the teacher provides answers. Therefore, comparing our datasets with the MathDial, QuAC, and NCTE datasets does not fully explain our datasets' quality, as MathDial, QuAC, and NCTE datasets are focused on different interaction situations. In particular, none of MathDial, QuAC, or NCTE datasets include textbook content; MathDial contains only short math problems and focuses on scaffolding, while QuAC is oriented towards fact-based queries rather than student-teacher interactions; NCTE consists of classroom transcripts in which there are more than just two interlocutors.

\section{Acknowledgements}
This project was made possible by an ETH AI Center Doctoral Fellowship to Jakub Macina with partial support from the Asuera Stiftung and the ETH Zurich Foundation. 
Nico Daheim acknowledges the funding by the German Federal Ministry of Education and Research and the Hessian
Ministry of Higher Education, Research, Science and the
Arts within their joint support of the National Research
Center for Applied Cybersecurity ATHENE.
We thank Xiaoyu Zhang for the insightful discussion about the design of the human evaluation experiments and Kumar Shridhar for help collecting the Openstax textbooks. Additionally, the authors wish to thank reviewers, members of the LRE group at ETH Zurich, and the participants in human evaluation experiments.

\section{Ethics and Broader Impact Statement}
We acknowledge the ethical implications and broader impacts of our work as follows:
\subsection{Ethical Considerations}
\paragraph{Data Privacy and Anonymity}
Our use of open-source textbooks from OpenStax ensures that the data is publicly available and free from privacy concerns. Additionally, in our human evaluation process, we rigorously removed all annotator information to maintain privacy and confidentiality.

\paragraph{Content Accuracy and Misinformation}
We recognize that our best data synthesis method has the problem of hallucinations, which may lead to misinformation. Continuous efforts to improve data accuracy and reduce misinformation are crucial.

\subsection{Broader Impacts}
\paragraph{Accessibility and Inclusivity}
By open-sourcing our data and code, we aim to enable a wider community to benefit from and contribute to this work.

\paragraph{Potential Misuse}
As with any AI application in the real-world, there is a potential for misuse. Our datasets and the accompanying code are intended to serve as supplementary resources in educational settings. It's important to emphasize that they should not replace human interactions and traditional teaching methods.

\bibliography{anthology,custom}

\appendix
\label{sec:appendix}

\section{Metrics Formulas} \label{sec:metric_formula}
The metrics mentioned in \Cref{sec:criteria} are detailed and explained in \Cref{tab:criterion_formula}, including formulas and explanations.

\begin{table*}[h!]
\centering
\small
\begin{tabular}{p{1.6cm}|p{1.5cm}|p{3.2cm}|p{7.9cm}}
\hline
\textbf{Criterion} & \textbf{Metric} & \textbf{Definition} & \textbf{Explanation}\\
\hline
\multirow{2}{*}{\specialcell{Answer\\Relevance}} & \specialcell{BF1\\($q_t$, $a_t$)} & BERTScoreF1($q_t$, $a_t$) &  For each QA pair, we compute the BERTScore F1\citep{bert-score}, treating the question as the predicted sentence and the answer as the reference sentence. It evaluates the semantic correspondence between the question and answer using BERT's contextual embeddings.\\
\cline{2-4}
 & QuestEval & QuestEval($q_t$, $a_t$) & For each QA pair, we compute the QuestEval score \citep{scialom-etal-2021-questeval}, treating the question as the predicted sentence and the answer as the reference sentence.  QuestEval generates questions from both the original question and the answer, then generates answers for these questions, comparing their consistency and completeness to evaluate Answer Relevance.\\
 \cline{2-4}
 & Uptake & Uptake($q_t$, $a_t$) & For each QA pair, we compute the Uptake score \citep{demszky-etal-2021-measuring} between student and teacher utterances. Uptake is computed as pointwise Jensen-Shannon Divergence (pJSD), estimated through next utterance classification, to analyze the teacher's responses to student utterances in terms of their dependence and relevance.\\
 \hline
 \multirow{2}{*}{\specialcell{Coherence}} & \specialcell{BF1\\($q_t$, $a_{<t}$)} & BERTScoreF1($q_t$, $a_{<t}$) & It computes the BERTScore F1 for each dialogue question, treating it as the predicted sentence against all preceding answers as references. Aggregated scores reflect the dialogue's coherence.\\
 \cline{2-4}
 &  \specialcell{BF1\\($q_t$, $a_{(t-1)}$)} & BERTScoreF1($q_t$, $a_{(t-1)}$) & It computes the BERTScore F1 for each dialogue question against the immediately preceding answer as the reference. Aggregated scores provide a measure of overall coherence.\\
 \hline

\specialcell{Informative-\\-ness} & \specialcell{1-Overlap\\($a_t$, $a_{<t}$)} & $ 1 - \frac{|\mathbf{a}_t \cap \mathbf{a}_{<t}|}{|\mathbf{a}_t \cup \mathbf{a}_{<t}|}$ &
For each answer in a dialogue, the proportion of its intersection with previous answers to their union is computed using word-level tokens. This value is then subtracted from 1.\\
 \hline
\multirow{2}{*}{\specialcell{Content\\Match}} & Density & $\frac{1}{|h_{1:T}|} \sum_{f \in \mathcal{F}(S,h_{1:T})} |f|^2$ ,$\mathcal{F}(S,h_{1:T})$:the set of extractive phrases in dialogue $h_{1:T}$ extracted from textbook content $S$.& Density refer to Extractive Fragment Density \citep{grusky-etal-2018-newsroom}, as the average length of text spans that are directly extracted from textbook content $S$ and included in the dialogues.  \\
\cline{2-4}
 & Coverage & $\frac{1}{|h_{1:T}|} \sum_{f \in \mathcal{F}(S,h_{1:T})} |f|$ & Coverage refer to Extractive Fragment Coverage \citep{grusky-etal-2018-newsroom}, as the percentage of words in a dialogue that originated from the textbook content.\\
 \hline
 Answerability & Answerable &  Valid(QA($q_t$,$S$)) &  We use the ``distilbert-base-cased-distilled-squad'' QA model to determine if a question is answerable from the textbook content. If it generates an empty string or an invalid answer such as ``CANNOTANSWER'', the question is deemed unanswerable. We report the ratio of answerable questions as 1 minus the ratio of unanswerable questions.\\
  \hline
\specialcell{Factual\\Consistency} & QFactScore & $\alpha sim(QA(q_t, S) , a_t)  + \beta sim(q_t, a_t)$ &
For each QA pair, it computes the cosine similarity between the embeddings of the QA model's predicted answer and the original answer. Then, it assesses the similarity between the embeddings of the question and answer. The final score is the weighted sum of two similarity scores.\\
 \hline
 Specificity & NA & NA & We lack automatic metrics for evaluating this criterion.\\
 
\hline
\end{tabular}
\caption{Formulas and Explanations of Each Metric: except 1-Overlap and QFactScore, all metrics are adopted from previous research.}
\label{tab:criterion_formula}
\end{table*}

\section{Textbook Statistics}

The four textbooks we used to generate dialogue for experiments were collected from the OpenStax website. The math textbook is titled 'Introductory Statistics,' the business textbook 'Business Ethics,' the science textbook 'Physics,' and the social science textbook 'Psychology 2e. The statistics of the four textbooks are shown in Table \ref{tab:textbook_statistics_4}.

\begin{table}[h!]
\centering
\scriptsize %
\setlength{\tabcolsep}{2pt} %
\begin{tabular}{l|p{1.5cm}|c|c|c|c}
\hline
 \textbf{Domain}  &
 \textbf{Name} &
 \textbf{Chapters} & \textbf{Paragraphs} & \textbf{Pages} & \textbf{Words} \\
\hline
Math  & \specialcell{Introductory\\Statistics} & 13 & 1,412 & 65 & 35,182\\
Business & \specialcell{Business\\Ethics} & 11  &  795 & 42 & 85,626\\
Science & Physics & 23 & 1,918 & 89 & 106,712 \\
Social science & Psychology 2e & 16 &  1,710 & 88 & 191,273\\
\hline
\end{tabular}
\caption{Summary of the textbook statistics: The Social Science textbook is the longest, while the Math textbook is the shortest.}
\label{tab:textbook_statistics_4}
\end{table}

\section{Implementation Details}\label{sec:textbook_to_dialogues_detail}
\subsection{Information-seeking scenario}
In the SimSeek-ASYM setup, the CQG model ingests the title and summary information, each separated by special tokens. We use T5-Large as the student's model and Longformer-Large as the teacher's model.

The SimSeek-ASYM code\footnote{https://github.com/naver-ai/simseek} can be executed with minor modifications. We use the same CQG and CAF models as in \citep{kim2022generating}, which utilize T5 as the student's model and Longformer as the teacher's model. 

\subsection{Dialogue Inpainting}\label{sec:dialogue_inpainting_detail}

We adopt a training regimen that integrates data from the OR-QuAC \citep{qu2020open}, QReCC \citep{anantha-etal-2021-open}, and the movie and restaurant datasets from Taskmaster-2 \citep{taskmaster2_2020}, employing the technique as described in \citep{dai2022dialog}. We randomly selected 80\% of the data as the training set, while the remaining 20\% as the test set. We implement Dialogue Inpainting using the code framework of \citep{daheim2023elastic}, basing our model (eq \ref{eq}) on FLAN-T5-XL \citep{chung2022scaling}, and train it with LoRA \citep{hu2022lora} to reduce computational load. We set an initial learning rate of 6.25e-5 and employed linear learning rate decay without warmup. For model optimization, we utilized checkpoints from the transformers library \citep{wolf2020transformers}. The negative log-likelihood of the ground-truth response was minimized using the AdamW optimizer, as detailed in \citep{loshchilov2019decoupled}. Model performance was assessed using the sacrebleu implementation of the BLEU metric, following \citep{post-2018-call}. We used one V100 GPU to train the model. The FLAN-T5-XL model has 3 billion parameters and took 12 hours to train.

The model, while fundamentally designed to predict single utterances, is used autoregressively. It begins with the input $s_{prompt}$, $<mask>$ , $s_1$ and sequentially generates questions using top-p sampling. This autoregressive process continues until the dialogue is wholly formed.

\subsection{Persona-based Generation}\label{sec:prompt}

\paragraph{Prompt for Persona (High Info)}
The design of our prompts was chiefly driven by the requisites of context-awareness, speaker identification, and specificity. We incorporated guidelines and annotations to ensure GPT yields concise responses and minimizes redundant information. To distinguish between speakers, we prefixed dialogues with labels:``Teacher:'' or ``Student:''. The prompt is shown below.

\textbf{Prompt for simulating student}

\noindent\texttt{
Task: You are a student preparing to ask questions about a textbook subsection to a teacher. 
Your goal is to uncover the key information from this subsection. Based on the teacher's responses, 
you'll further inquire to get a comprehensive understanding. Make sure to ask specific questions about 
the subsection's content and avoid repeating queries from prior discussions.}

\noindent\texttt{Information Provided:
\begin{enumerate}
    \item Section Title: ...
    \item Subsection Title: ...
    \item Section Summary: ...
    \item Bold Terms in Section: ...
    \item Learning Objectives: ...
    \item Concepts in Section: ...
    \item Section Introduction: ...
\end{enumerate}
Previous Conversation:
\begin{itemize}
    \item[] Student:...
    \item[] Teacher:...
\end{itemize}}

\noindent\texttt{*Note:* Frame your questions considering the information above and ensure they're relevant to the content. Do not ask questions about information you already have. Only ask one question at a time.}

\noindent\texttt{Expected Output: Please phrase your question as a string.}

\paragraph{Prompt for simulating teacher}

\noindent\texttt{Task: You are a teacher preparing to answer a student's question about a subsection of a textbook. The student's question is: \{question\}. Provide a concise, specific response, ensuring it's not a summary and distinct from any previous answers you've given.}

\noindent\texttt{Information Provided:
\begin{enumerate}
    \item Section Title: ...
    \item Subsection Title: ...
    \item Subsection Content: ...
    \item Section Summary: ...
    \item Bold Terms in Section: ...
    \item Learning Objectives: ...
    \item Concepts in Section: ...
    \item Section Introduction: ...
\end{enumerate}
Previous Conversation: 
\begin{itemize}
    \item[] Student:...
    \item[] Teacher:...
\end{itemize}}

\noindent\texttt{*Note:* When crafting your response, consider all the information above. Be sure your answer directly addresses the student's question and is not a repetition of prior information.}

\noindent\texttt{Expected Output: Please phrase your answer as a string.}

\paragraph{Prompt for Persona (Single Instance)}
The prompt for the Persona (Single Instance) method is shown below. It uses one prompt to generate one dialogue.

\noindent\texttt{Task: generate a conversation between a student and a teacher using the given section.}
    
\noindent\texttt{Introduction:
    \begin{enumerate}
    \item  The conversation should contain 6 question-answer pairs.
    \item  The output conversation should be in this format: student: ... teacher: ... student: ...
    \item  The given section: ...
    \end{enumerate}}

\subsection{QFactScore Implementation}\label{sec:qfactscore}
For computing the embeddings of questions and answers, we use the ``msmarco-distilbert-cos-v5'' model from \citep{reimers-gurevych-2019-sentence}. This model is suitable for computing cosine similarity and performs well in our task.

It is important to ensure that the QA model used in QFactScore is different from the QA model used for generating dialogue datasets. This is because if the same QA model is used, the predicted answer is likely to be similar to the original answer in the dialogue. In QFactScore, we use the 'distilbert-base-cased-distilled-squad' model, which differs from the GPT-3.5, T5, and Flan-T5 models that we used for generating the dataset.

QFactScore computes as the below equation. For each QA pair, it computes the cosine similarity between the embeddings of the QA model's predicted answer and the original answer. Then, it assesses the similarity between the embeddings of the question and answer. The final score is the weighted sum of two similarity scores. The weight can be adjusted according to different applications, in our study we use $\alpha = 1$ and $\beta = 1$.

\begin{equation}
\alpha\cdot\text{sim}(\text{QA}(q_t, S) , a_t)  + \beta\cdot\text{sim}(q_t, a_t)
\end{equation}

We further evaluate the correlation between QFactScore and human evaluation of Factual Consistency in \Cref{sec:metric_eval}. We also provide correlation between 1 - Overlap($a_t$,$a_{<t}$) and human evaluation of Informativeness in \Cref{sec:metric_eval}.

\section{Methods Comparison}
The details of different types of Student's Models and Teacher's Models of each generation method and the detail of corresponding input are listed in Table \ref{tab:model_comparison}. The term ``Formatting'' refers to formatting information, which contains a title, summary, introduction, learning objectives, bold terms, and key concepts from textbooks, which is introduced in Section-\ref{sec:task}. The ``COPY'' in the teacher's model of Dialogue Inpainting indicates this method just copying a sentence from the textbook as the answer.

\begin{table*}[h!]
\centering
\setlength{\tabcolsep}{2pt} %
\begin{tabular}{l|c|c|p{3cm}|p{3cm}}
\hline
 \textbf{Methods} &  \textbf{Student's Model} &  \textbf{Teacher's Model} & \textbf{Inpute to Student} & \textbf{Input to Teacher}
 \\
\hline
SimSeek & T5 & Longformer & Title + Summary & \multirow{6}{*}{Contents+Formatting} \\
Dialog Inpainting & FLAN-T5 & COPY & Contents + Formatting &  \\
\cline{1-4}
Persona (Low Info) & \multirow{4}{*}{GPT-3.5} & \multirow{4}{*}{GPT-3.5} & Title & \\
Persona (Medium Info) &  &  & Title + Summary & \\
Persona (High Info) &  &  & Formatting &  \\
Persona (Single Instance) &  &  & Contents + Formatting &  \\
\hline
\end{tabular}
\caption{Summary of methods used in this work showing details of different types of student models and teacher models of each generation method and the detail of corresponding input.}
\label{tab:model_comparison}
\end{table*}

\section{Metric Evaluation}\label{sec:metric_eval}
To validate the effectiveness of the metrics introduced in this study, we calculated both Pearson and Spearman correlations between the metrics' outcomes and the corresponding results from human evaluations. The results are shown in Table \ref{tab:key_pearson_result} and Table \ref{tab:key_spearman_result}. The ``1 - Overlap($a_t$,$a_{<t}$)'' exhibits a Pearson correlation of 0.81 and a Spearman correlation of 0.77 with the Informativeness score in human evaluation, both with p-values below 0.005, suggesting that this F1 score could effectively represent Informativeness in evaluations.

On the other hand, QFactScore exhibits a Pearson correlation of 0.35 and a Spearman correlation of 0.38 with Factual Consistency in human evaluation, both with p-values below 0.005. We interpret this as indicative of a moderate correlation, suggesting that this metric can approximate factual consistency to a certain extent. When comparing the correlation results with existing methods, including the use of GPT-3.5 scores derived from prompts, QuestEval, and QrelScore, the findings indicate that QFactScore's correlation score surpasses others. However, Factual Consistency is a nuanced criterion that necessitates an assessment of whether the answer accurately addresses the question within the given context. Existing metrics struggle with this task, highlighting the need for more comprehensive evaluations in the future.

\begin{table}[h!]
\centering
\small %
\setlength{\tabcolsep}{2pt} %
\begin{tabular}{lcccc}
\hline
 & & & Correlation & P-Value
 \\
\hline
\specialcell{1 - Overlap($a_t$,$a_{<t}$)} & vs & \specialcell{Informativeness} & \textbf{0.81} & \textbf{0.002} \\
\specialcell{1 - BF1($a_t$,$a_{<t}$)} & vs & \specialcell{Informativeness} & 0.69 & 0.01
 \\
\specialcell{QFactScore} & vs & \specialcell{Factual\\Consistency} &  \textbf{0.35} & \textbf{0.003} \\
\specialcell{GPT-3.5} & vs & \specialcell{Factual\\Consistency} &  0.28 & 0.02 \\
\specialcell{QuestEval} & vs & \specialcell{Factual\\Consistency} &  0.30 & 0.01 \\
\specialcell{QrelScore} & vs & \specialcell{Factual\\Consistency} &  -0.035 & 0.77 \\
\hline
\end{tabular}
\caption{Pearson correlation of metrics and human evaluation: indicates that the $1-Overlap$ metric has good alignment with human judgment, while QFactScore shows moderate alignment.}
\label{tab:key_pearson_result}
\end{table}

\begin{table}[h!]
\centering
\small %
\setlength{\tabcolsep}{2pt} %
\begin{tabular}{lcccc}
\hline
 & & & Correlation & P-Value
 \\
\hline
 \specialcell{1 - Overlap($a_t$,$a_{<t}$)} & vs & \specialcell{Informativeness} & \textbf{0.77} & \textbf{0.0038} \\
  \specialcell{1 - BF1($a_t$,$a_{<t}$)} & vs & \specialcell{Informativeness} & 0.76 & 0.0040
 \\
\specialcell{QFactScore} & vs & \specialcell{Factual\\Consistency} &  \textbf{0.38} & \textbf{0.0009} \\
\specialcell{GPT-3.5} & vs & \specialcell{Factual\\Consistency} &  0.29 & 0.01 \\
\specialcell{QuestEval} & vs & \specialcell{Factual\\Consistency} &  0.33 & 0.004 \\
\specialcell{QrelScore} & vs & \specialcell{Factual\\Consistency} &  0.08 & 0.51 \\
\hline
\end{tabular}
\caption{Spearmans correlation of metrics and human evaluation: indicates that the $1-Overlap$ metric has good alignment with human judgment, while QFactScore shows moderate alignment.}
\label{tab:key_spearman_result}
\end{table}

\subsection{Metrics Results Details}
We provide the complete results of different metrics for datasets in four domains in this section. The results are shown in Table \ref{tab:statistic_table_full} and Table \ref{tab:metric_table_full}.
\begin{table*}
\centering
\small
\begin{tabular}{l|l|ccc|cc}
\hline
\multirow{2}{*}{\textbf{Domain}} & \multirow{2}{*}{\textbf{Models}} & \multicolumn{3}{c|}{\textbf{ Question Type}}  & \multicolumn{2}{c}{\textbf{Number of Tokens}}\\
\cline{3-7}
& & \textbf{\tiny \specialcell{ \%``what'' or ``which''}} & \textbf{\tiny \specialcell{ \%``why''}} & \textbf{\tiny \specialcell{ \%``how''}}  & \textbf{\tiny \specialcell{ Avg Tokens\\in Questions}} & \textbf{\tiny \specialcell{ Avg Tokens\\in Answers}}\\
\hline
\multirow{6}{*}{Math} & SimSeek & 52 & 2 & 20 & 11.24 & 11.66 \\
& Dialogue Inpainting & 60 & 2 & 17 & 7.55 & 15.41\\
& Persona (Low Info) & 62 & 1 & 31  & 18.19 & 80.16\\
& Persona (Medium Info) & 64 & 0 & 29  & 18.28 & 81.83\\
& Persona (High Info) & 76 & 1 & 23  & 19.95 & 77.96 \\
& Persona (Single Instance) & 32 & 5 & 27  & 15.57 & 29.96 \\
\hline
\multirow{6}{*}{Business} & SimSeek & 51 & 2 & 18  & 11.19 & 16.17 \\
& Dialogue Inpainting & 66 & 3 & 14  & 6.75 & 23.79\\
& Persona (Low Info) & 75 & 0 & 24  & 17.74 & 99.03\\
& Persona (Medium Info) & 70 & 0 & 26  & 18.94 & 99.36\\
& Persona (High Info) & 76 & 0 & 22  & 19.52 & 98.86\\
& Persona (Single Instance) & 24 & 8 & 21  & 16.28 & 40.84 \\
\hline
\multirow{6}{*}{Science} & SimSeek& 59 & 2 & 17  & 10.73 & 14.73\\
& Dialogue Inpainting & 62 & 3 & 18  & 6.55 & 17.91 \\
& Persona (Low Info) & 71 & 1 & 32  & 17.50 & 83.12\\
& Persona (Medium Info) & 69 & 0 & 32  & 16.71 & 83.70\\
& Persona (High Info) & 71 & 1 & 28  & 18.40 & 84.05\\
& Persona (Single Instance) & 30 & 5 & 25  & 13.43 & 31.31 \\
\hline
\multirow{6}{*}{\specialcell{Social\\Science}} & SimSeek & 58 & 2 & 15  & 10.42 & 14.74 \\
& Dialogue Inpainting & 66 & 3 & 15 & 6.53 & 21.42 \\
& Persona (Low Info) & 73 & 1 & 23  & 16.82 & 76.68\\
& Persona (Medium Info) & 73 & 1 & 21  & 16.82 & 75.87\\
& Persona (High Info) & 71 & 1 & 23  & 18.15 &  77.91\\
& Persona (Single Instance) & 27 & 4 & 20 & 14.58 &  36.20 \\
\hline
\end{tabular}
\caption{
Dataset statistics in more detail: Persona-generated datasets contain more 'what/which' and 'how' questions, and the dialogues tend to be more verbose.
}
\label{tab:statistic_table_full}
\end{table*}

\begin{table*}
\centering
\resizebox{\linewidth}{!}{
\begin{tabular}{l|l|ccc|c|cc|cc|c|c}
\hline
\multirow{2}{*}{\textbf{Domain}} & \multirow{2}{*}{\textbf{Models}} & \multicolumn{3}{c|}{\textbf{\specialcell{Answer\\Relevance}}} & \multicolumn{1}{c|}{\textbf{\specialcell{Informativeness}}} & \multicolumn{2}{c|}{\textbf{\specialcell{Groundedness}}} & \multicolumn{2}{c|}{\textbf{\specialcell{Coherence}}}  & \textbf{Answerability}& \multicolumn{1}{c}{\textbf{\specialcell{Factual\\Consistency}}} \\
\cline{3-12}
& & \textbf{ \specialcell{BF1\\($q_t$, $a_t$)}} & \textbf{ QuestEval} & \textbf{ Uptake} & \textbf{ \specialcell{1 - Overlap\\($a_t$, $a_{<t}$)}}  & \textbf{ Density} & \textbf{ Coverage} & \textbf{ \specialcell{BF1\\($q_t$,  $a_{<t}$)}}  & \textbf{\specialcell{BF1\\($q_t$, $a_{t-1}$)}} & \textbf{ Answerable} & \textbf{ QFactScore} \\
\hline
\multirow{7}{*}{Math} & SimSeek & 0.51 & 0.24 & 0.64 & 0.61  & 9.5 & 0.71 & 0.49 & 0.53 & 0.74 & 0.27  \\
& \specialcell{Dialogue\\Inpainting} & 0.57 & 0.30 & 0.84 & \textbf{0.88}  & \textbf{19.37} & \textbf{0.88}  & 0.46 & 0.47 & 0.52 & 0.19\\
& \specialcell{Persona\\(Single Instance)} & 0.58 & 0.32 & 0.97 & 0.85 & 2.94 & 0.62 & 0.50 & 0.52 & 0.87 & 0.53\\
& \specialcell{Persona\\(Low Info)} & \textbf{0.62} & \textbf{0.43} & \textbf{0.99} & 0.54  &  1.94 & 0.59  & 0.51 & 0.59 & 0.99 & 0.80\\
& \specialcell{Persona\\(Medium Info)} & \textbf{0.62} & \textbf{0.43} & \textbf{0.99} & 0.55  & 2.09 & 0.60 & 0.51 & 0.59 & \textbf{1.00} & \textbf{0.81}\\
& \specialcell{Persona\\(High Info)} & \textbf{0.62} & \textbf{0.43} & \textbf{0.99} & 0.56 & 2.07 & 0.60  & \textbf{0.52} & \textbf{0.60} & 0.99 & \textbf{0.81}\\

\hline
\multirow{7}{*}{Business} & SimSeek & 0.54 & 0.25 & 0.82 & 0.77  & 13.16 & 0.88  & 0.52 & 0.56 & 0.89 & 0.32 \\
& \specialcell{Dialogue\\Inpainting} & 0.49  & 0.26 & 0.81 & \textbf{0.94}  & \textbf{26.44} & \textbf{0.92}  & 0.43 & 0.45 & 0.88 & 0.23 \\
& \specialcell{Persona\\(Single Instance)} & 0.58 & 0.36 & \textbf{0.99} & 0.88 & 4.07 & 0.82  & 0.50 & 0.53 & 0.95 & 0.52 \\
& \specialcell{Persona\\(Low Info)} & 0.62 & \textbf{0.46} & \textbf{0.99} & 0.61  & 2.38 & 0.76  & 0.52 & \textbf{0.60} & 0.99 & 0.73\\
& \specialcell{Persona\\(Medium Info)} & 0.62 & \textbf{0.46} & \textbf{0.99} & 0.61 & 2.31 & 0.77  & 0.53 & \textbf{0.60} & \textbf{1.00} & 0.73 \\
& \specialcell{Persona\\(High Info)} & \textbf{0.63} & \textbf{0.46} & \textbf{0.99} & 0.62 & 2.44 & 0.77   & \textbf{0.54} & \textbf{0.60} & \textbf{1.00} &\textbf{0.74}\\

\hline
\multirow{7}{*}{Science} & SimSeek & 0.52 & 0.25 & 0.81 & 0.71  & 11.78 &  0.83  & 0.50 & 0.54 & 0.89 & 0.34 \\
& \specialcell{Dialogue\\Inpainting} & 0.51 & 0.27 & 0.82 & \textbf{0.92} & \textbf{20.43} & \textbf{0.90}  & 0.44 & 0.44 & 0.72 & 0.24\\
& \specialcell{Persona\\(Single Instance)} & 0.58 & 0.35 & 0.98 & 0.85 & 4.65 & 0.79  & 0.48 & 0.51 & 0.94 & 0.61\\
& \specialcell{Persona\\(Low Info)} & \textbf{0.59} & \textbf{0.43} & \textbf{0.99} & 0.57  & 2.55 & 0.73    & \textbf{0.51} & \textbf{0.57} & 0.98 & 0.79\\
& \specialcell{Persona\\(Medium Info)} & \textbf{0.59} & \textbf{0.43} & \textbf{0.99} & 0.57  & 2.63 & 0.73   & 0.50 & \textbf{0.57} & \textbf{0.99} & \textbf{0.80}\\
& \specialcell{Persona\\(High Info)} & \textbf{0.59}  & \textbf{0.43} & \textbf{0.99} & 0.58  & 2.68 & 0.74   & \textbf{0.51} & \textbf{0.57} & \textbf{0.99} & 0.76\\

\hline
\multirow{7}{*}{\specialcell{Social\\Science}} & SimSeek & 0.53 & 0.27 & 0.85 & 0.74  & 12.21 & 0.84  & 0.51 & 0.55 & 0.89 & 0.34\\
& \specialcell{Dialogue\\Inpainting} & 0.51 & 0.28 & 0.87 & \textbf{0.91}  & \textbf{24.22} & \textbf{0.91}   & 0.45 & 0.48 & 0.86 & 0.29\\
& \specialcell{Persona\\(Single Instance)} & 0.57 & 0.36  & \textbf{0.99}  & 0.87 & 4.09 & 0.77  & 0.49 & 0.52 & 0.92 & 0.50 \\
& \specialcell{Persona\\(Low Info)} & \textbf{0.62} & \textbf{0.45}  & \textbf{0.99}  & 0.63  & 2.67 & 0.73  & 0.52 & 0.59 & 0.98 & 0.69\\
& \specialcell{Persona\\(Medium Info)} & \textbf{0.62} & \textbf{0.45}  & \textbf{0.99}  & 0.64  & 2.69 & 0.73   & 0.52 & \textbf{0.60} & \textbf{1.00} & \textbf{0.71}\\
& \specialcell{Persona\\(High Info)} & \textbf{0.62} & \textbf{0.45}  & \textbf{0.99}  & 0.63  & 2.79 & 0.74   & \textbf{0.53} & 0.59 & 0.99 & 0.69\\

\hline
\end{tabular}
}
\caption{
Metrics results break down for datasets in different domains: the bolded value represents the best value for each criterion within each domain. Higher values indicate better performance. Persona-generated dialogues score highest in Answer Relevance, Coherence, Answerability, and Factual Consistency, while Dialogue Inpainting generated dialogues score highly in Informativeness and Groundedness.
}
\label{tab:metric_table_full}
\end{table*}

\section{Human Evaluation Details}\label{sec:human_eval_detail}
\subsection{Experiment Details}
We have adopted a human evaluation approach to assess the performance of dialogues generated by various methods. We recruited four annotators who have master's degrees in Math, Science, Social Science, and Business. The annotators have educational backgrounds in Europe and Asia and are aged between 20 and 25. We recruit them by advertising on social media and reward them with non-monetary gifts for each annotator. As all annotators are satisfied with this payment, we consider this as adequate. To alleviate the burden on participants, we selected the 3 models from each method category for evaluation. To ensure the consistency of results across different domains, we chose datasets from four textbooks, each covering a different subject area: mathematics, business, science, and social sciences. From each textbook, we randomly selected a subsection. For each subsection, we generated one dialogue using a different method, preparing each dialogue separately for evaluation. We use only the first 12 turns (6 QA pairs) of each dialogue for evaluation, similar to what is described in Section \ref{sec:metric_result}. During the evaluation, each of the three participants received 12 dialogues, with every dialogue corresponding to a related textbook subsection. Evaluators rated each question-answer (QA) pair within a dialogue based on eight criteria. The overall evaluation score for a dialogue was determined by averaging the scores of all its QA pairs. The specific evaluation criterion and corresponding questions are detailed in Table \ref{tab:human_eval_criterion}. Participants responded to each question with ``yes'' or ``no''. The ``yes'' is coded as a score of 1, while the ``no'' is coded as a score of 0.

We provide the specific question the participants will be asked during human evaluation as shown in Table \ref{tab:human_eval_criterion}. The task is straightforward, we provide QA pairs for evaluation in an Excel file and the annotators just read the QA pair and give a score based on their judgement of each question. 

We further show the Cohen's Kappa score between each participant in Table \ref{tab:cohen}, which proves that each pair of participants has substantial agreement.
\begin{table}[h!]
\centering
\small
\begin{tabular}{p{2cm}|p{5cm}} %
\hline
\textbf{Criterion} & \textbf{Questions for each QA pair} \\
\hline
\specialcell{Answer\\Relevance} & \textbf{Question}: Is the response directly addressing the posed question? (answer no if it is answering a different question)\\
\hline
\makecell[l]{Informativeness} & \textbf{Question}: Does the current answer introduce new information that was not mentioned in previous answers within the same conversation? \\
\hline
\makecell[l]{Groundedness} & \textbf{Question}: Does the answer contain specific details or data points mentioned in the contextual background or previous dialogue?\\
\hline
\makecell[l]{Coherence} & \textbf{Question 1}: Does the current question directly follow up on the immediate previous answer? (Ignore the first QA pair) \\
\hline
Factual Consistency & \textbf{Question}: Does the answer correctly address the question, considering the context provided? (If 'answerability' is 'no,' then this criterion should also be 'no.')\\
\hline
Answerability & \textbf{Question}: Can the question be answered given the context?\\
\hline
Specificity & \textbf{Question}: Does this question exhibit generality, such that it could be relevant beyond the immediate context provided? (e.g. What is interesting about this passage?)\\
\hline
\end{tabular}
\caption{Exact framing of questions asked during the human evaluation: annotators answer each question with yes or no for each QA pair. }
\label{tab:human_eval_criterion}
\end{table}

\begin{table}
\centering
\normalsize
\begin{tabular}{l|c}
\hline
\textbf{Participants Pairs} & \textbf{Cohen's Kappa} \\
\hline
P1 vs. P2 & 0.71 \\
P1 vs. P3 & 0.71 \\
P1 vs. P4 & 0.90 \\
P2 vs. P3 & 0.73 \\
P2 vs. P4 & 0.71 \\
P3 vs. P4 & 0.71 \\
\hline
\end{tabular}
\caption{
The Cohen's Kappa score between each pair of participants: indicates reasonable alignment between annotators.
}
\label{tab:cohen}
\end{table}

\subsection{Disclaimer for Annotators}
Thank you for participating in our evaluation process. Please read the following important points before you begin:

\begin{itemize}
    \item \textbf{Voluntary Participation:} Your participation is completely voluntary. You have the freedom to withdraw from the task at any time without any consequences.
    \item \textbf{Confidentiality:} All data you will be working with is anonymized and does not contain any personal information. Your responses and scores will also be kept confidential.
    \item \textbf{Risk Disclaimer:} This task does not involve any significant risks. It primarily consists of reading and scoring QA pairs.
    \item \textbf{Queries:} If you have any questions or concerns during the task, please feel free to reach out to us.
\end{itemize}

\subsection{Instructions for Experiments}

Thank you for participating in our evaluation experiment. The data collected through this process will be used to assess the quality of our methods. 

Follow these steps to score each QA pair:

\begin{enumerate}
    \item \textbf{Accessing the Data:} Open the provided Excel file, which contains the QA pairs for evaluation.
    \item \textbf{Scoring Each QA Pair:} For each pair, read the question and the corresponding answer carefully.
    \item \textbf{Scoring Scale:} Answer each question with ``yes'' or ``no''.
    \item \textbf{Entering Scores:} Enter your score for each QA pair in the designated column in the Excel sheet. Please stick to the scoring scale provided.
    \item \textbf{Consistency:} Try to maintain consistency in your scoring. Refer to the example evaluations provided if you're unsure.
    \item \textbf{Completion:} Once you have scored all the QA pairs, save the file and return it to us as instructed.
\end{enumerate}

We appreciate your time and effort in this task.

\section{Pre-training for Educational Chatbots Details}\label{sec:downstream_task_detail}
We sourced four textbooks from the OpenStax website for our study. These include 'Introductory Statistics' for math, 'Business Ethics' for business studies, 'Physics' for science, and 'Psychology 2e' for social science. We use the entire textbook dialogue dataset for pre-training.

In line with the methodology described in \citep{macina-etal-2023-mathdial}, the models with pre-train were trained 10 epochs during pre-train and trained 10 epochs during fine-tune. The models without pre-train trained 10 epochs during training. For CoQA CNN and MCTest dialogue datasets for fine-tune or training, we use 60\% of data for training, 20\% for validation, and 20\% for testing. For the NCTE dataset, we randomly select 10,000 dialogues for training, 2,000 dialogues for validation, and 2,000 dialogues for testing. We set an initial learning rate of 6.25e-5 and employed linear learning rate decay without warmup. For model optimization, we utilized checkpoints from the transformers library \citep{wolf2020transformers}. The negative log-likelihood of the ground-truth response was minimized using the AdamW optimizer, as detailed in \citep{loshchilov2019decoupled}. Model performance was assessed using the sacrebleu implementation of the BLEU metric, following \citep{post-2018-call}. We used one V100 GPU to train the model, the FLAN-T5-LARGE model has 0.8 Billion parameters and took 7 hours to train. The result in \Cref{tab:chatbot} is the average of 3 runs.

\subsection{Examples of Results}\label{sub:Downstream_Task_Generation_Results}
We demonstrate an example of a comparison between the ground truth, the prediction from the model without pretraining, and the prediction from the model with pretraining in \Cref{tab:coqa_mctest_comparison}. These dialogues are based on the MCTest split of the CoQA dataset and the predictions are generated based on the same context information as shown in \Cref{tab:coqa_mctest_context}. From \Cref{tab:coqa_mctest_comparison}, we find that in the third answer, the prediction example without fine-tuning is overly verbose, suggesting a poor grasp of the question, 'What was the \textbf{first thing} she did that morning?' which focuses on the person's initial action. In contrast, the model without fine-tuning simply replicates the entire context sentence. On the other hand, the prediction with fine-tuning accurately mirrors the ground truth dialogue, effectively comprehending the concept of the \textbf{"first thing"} and suitably condensing the sentence from the context.

\begin{table}[h!]
\centering
\small %
\setlength{\tabcolsep}{2pt} %
\begin{tabular}{p{7cm}}
When Sophie woke up that morning, she had no idea where her day was going to take her. She rolled out of bed, turned off her alarm and stretched. She wasn't feeling like herself that morning, but she wasn't sure why. Sophie thought to herself, "I slept well, I ate good food yesterday, and yet I still felt strange". Sophie stepped into the shower feeling so tired. As she toweled herself off and got dressed, she felt like she was moving very slowly. She went to the kitchen and poured herself a glass of orange juice, got a bowl out of the cabinet and filled it with cereal. As Sophie sat at the table to eat her breakfast, she remembered why she wasn't feeling like herself. She remembered that she didn't exercise the day before. She had spent much of the day sitting in front of her television and playing games. Sophie knew that on days that she didn't exercise, she always felt bad the next day. Sophie thought for a second, and then looked at the clock. She had time to do some jumping jacks and run outside around her house before she had to leave for the day. She put on her running shoes and went out the front door. After only a quick bit of exercise, Sophie was feeling much better. She promised herself that she would never forget to exercise again.

\end{tabular}
\caption{Context of the example CoQA MCTest split dialogue: the context provides grounded information for the generated dialogues.}
\label{tab:coqa_mctest_context}
\end{table}

\begin{table*}[h!]
\centering
\small %
\setlength{\tabcolsep}{2pt} %
\begin{tabular}{p{5cm}|p{5cm}|p{5cm}}
Ground Truth & Prediction without Fine-tuning & Prediction with Fine-tuning \\
\hline
 \begin{tabular}{@{}p{5cm}@{}}
\textbf{Q:} Who is waking up?\\
\textbf{A:} Sophie\\
\textbf{Q:} How was she feeling today?\\
\textbf{A:} Tired. \\
\textbf{Q:} What was the first thing she did that morning?\\
\textbf{A:} She rolled out of bed. \\
\textbf{Q:} What woke her up that morning? \\
\textbf{A:} Her alarm \\
\textbf{Q:} What did she do after turning the alarm off? \\
\textbf{A:} Stretched. \\
\textbf{Q:} What did she do next? \\
\textbf{A:} Showered. \\
\end{tabular} 
&
\begin{tabular}{@{}p{5cm}@{}}
\textbf{Q:} Who is waking up?\\
\textbf{A:} Sophie\\
\textbf{Q:} How was she feeling today?\\
\textbf{A:} Tired. \\
\textbf{Q:} What was the first thing she did that morning?\\
\textbf{A:} \textbf{She rolled out of bed, turned off her alarm and stretched.} \\
\textbf{Q:} What woke her up that morning? \\
\textbf{A:} Her alarm \\
\textbf{Q:} What did she do after turning the alarm off? \\
\textbf{A:} Stretched. \\
\textbf{Q:} What did she do next? \\
\textbf{A:} Showered. \\
\end{tabular} 
&
\begin{tabular}{@{}p{5cm}@{}}
\textbf{Q:} Who is waking up?\\
\textbf{A:} Sophie\\
\textbf{Q:} How was she feeling today?\\
\textbf{A:} Tired. \\
\textbf{Q:} What was the first thing she did that morning?\\
\textbf{A:} She rolled out of bed. \\
\textbf{Q:} What woke her up that morning? \\
\textbf{A:} Her alarm \\
\textbf{Q:} What did she do after turning the alarm off? \\
\textbf{A:} Stretched. \\
\textbf{Q:} What did she do next? \\
\textbf{A:} Showered. \\
\end{tabular}

\end{tabular}
\caption{CoQA MCTest split dialogue comparison: The prediction example without fine-tuning provides verbose content in the third answer, indicating that the model does not have a good understanding of the question which asks, 'What was the \textbf{first thing} she did that morning?', emphasizing the initial action of the person. However, the model without fine-tuning returns the entire sentence from the context. Meanwhile, the prediction with fine-tuning shows an exact match with the ground truth dialogue, demonstrating that the model can understand the concept of the \textbf{first thing} and appropriately truncate the sentence from the context.}
\label{tab:coqa_mctest_comparison}
\end{table*}

\subsection{Generated Datasets Overview}
We provide the overview of our generated dataset in \Cref{tab:datasets_overview}.
\begin{table*}[h!]
\centering
\small %
\begin{tabular}{l|l|c|c|c|c}
\hline
 Domain & Generation Method & Dialogues & Dialogic Pairs & Bigram Entropy & Avg. words per utterance \\
\hline
\multirow{2}{*}{Math} & Persona (High Info) & 142 & 852 & 6.08 & 48.95 \\
& Dialog Inpainting & 142 & 1444 & 4.07 & 11.05 \\
\hline
\multirow{2}{*}{Business} & Persona (High Info) & 123 & 738 & 6.61 & 59.01 \\
& Dialog Inpainting & 123 & 3575 & 4.46 & 14.39 \\

\hline
\multirow{2}{*}{Science} & Persona (High Info) & 228 & 1368 & 6.22 & 48.03 \\
& Dialog Inpainting & 228 & 5898 & 4.56 & 13.99 \\
\hline
\multirow{2}{*}{Social} & Persona (High Info) & 396 & 2376 & 6.2 & 51.04 \\
& Dialog Inpainting & 396 & 7503 & 4.34 & 11.69 \\
\hline
\multicolumn{2}{c|}{Total} & 1778 & 23754 & 5.3175 & 19.48875 \\
\hline
\end{tabular}
\caption{Detailed overview of the synthetic dataset: the Persona model generated dialogues are more verbose and diverse than dialogues generated by the dialogue inpainting model.}
\label{tab:datasets_overview}
\end{table*}

\subsection{Implementation Details}

For each educational chatbot, we used the code framework of \citet{daheim2023elastic} with an initial learning rate of 6.25e-5 and linear learning rate decay without warmup. For model optimization, we utilized checkpoints from the transformers library \citep{wolf2020transformers}. The negative log-likelihood of the ground-truth response was minimized using the AdamW optimizer, as detailed in \citep{loshchilov2019decoupled}. Model performance was assessed using the sacrebleu\footnote{https://pypi.org/project/sacrebleu/} implementation of the BLEU metric, following \citep{post-2018-call}. We used one V100 GPU to train the model. The FLAN-T5-LARGE model has 0.8 billion parameters and took 7 hours to train.

\subsection{Details of Insufficient Follow-up Ability of the System}
To evaluate the follow-up ability of the system, we conducted a random examination of 12 dialogues of each model. Our analysis focused on the continuity and depth of the QA pairs within each dialogue. Although the pairs consistently discuss the same topic, we observed a lack of in-depth follow-up on previously mentioned aspects. For instance, instead of a pattern where an initial dialogue introduces a piece of knowledge and subsequent dialogues delve deeper into that same knowledge, the conversations often shift to other related aspects. We present an example of a dialogue demonstrating insufficient follow-up ability of the Persona-based model  in \cref{tab:dialogue_example_insufficient}.

\begin{table}
\centering
\small %
\setlength{\tabcolsep}{2pt} %
\begin{tabular}{p{7cm}}
\textbf{Student:} What are \textbf{monocular} cues and how do they contribute to our perception of depth in the context of the section "5-3-vision"? \\
\textbf{Teacher:} Monocular cues are visual cues that only require one eye and contribute to our perception of depth. In the context of the section "5-3-vision\ monocular cues such as linear perspective, interposition, and relative size and closeness of images help us perceive depth in 2-D images, such as paintings and photographs. These cues allow us to understand the spatial relationships between objects in a scene and create a sense of depth, despite the visual stimulus being 2-D. \\
\textbf{Student:} How do \textbf{binocular} cues like linear perspective and interposition contribute to our perception of depth in the context of the section "5-3-vision"? \\
\textbf{Teacher:} Binocular cues like linear perspective and interposition contribute to our perception of depth in the context of the section "5-3-vision" by providing additional visual information to our brain through the use of both eyes. Linear perspective refers to the perception of depth when we see two parallel lines appearing to converge, while interposition refers to the partial overlap of objects. These binocular cues, along with other cues like binocular disparity, help us establish a sense of depth and understand the spatial relationships between objects in a scene, both in our three-dimensional world and in 2-D images such as paintings and photographs. \\
\textbf{Student:} How do \textbf{monocular} cues like linear perspective, interposition, and relative size and closeness of images contribute to our perception of depth in the context of the section "5-3-vision"? \\
\textbf{Teacher:} Linear perspective and interposition, as binocular cues, contribute to our perception of depth by providing additional visual information through the use of both eyes. Linear perspective refers to the perception of depth when we see two parallel lines appearing to converge, while interposition refers to the partial overlap of objects. These cues, along with binocular disparity and other monocular cues, help us establish a sense of depth and understand the spatial relationships between objects in our three-dimensional world and in 2-D images such as paintings and photographs. \\
\end{tabular}
\caption{Example of Dialogue with Insufficient Follow-up Ability: This table illustrates a dialogue sequence where the first QA pair addresses monocular cues, but the following pair shifts focus to binocular cues without maintaining topical continuity. The dialogue then reverts to monocular cues in the third pair, demonstrating an overall lack of follow-up ability in the dialogue. For each teacher's response, the student also does not give any feedback and just jumps to the next question, which also does not align with natural conversations.}
\label{tab:dialogue_example_insufficient}
\end{table}

\subsection{Example of Generated Dialogues}
We show examples of generated dialogues from each model in this section. Example dialogues generated are shown in \Cref{tab:dialogue_example_all}, all these dialogues are generated based on the same context and formatting information as shown in \cref{tab:context_formatting}.

\begin{table*}[h!]
\centering
\setlength{\tabcolsep}{2pt} %
\begin{tabular}{p{7cm}|p{5cm}}
\textbf{Context} & \textbf{Formatting} \\
\hline

 \begin{tabular}{@{}p{7cm}@{}}
   We learned in the previous chapter that when two objects (or systems) are in contact with one another, heat will transfer thermal energy from the object at higher temperature to the one at lower temperature until they both reach the same temperature. The objects are then in thermal equilibrium, and no further temperature changes will occur if they are isolated from other systems. The systems interact and change because their temperatures are different, and the changes stop once their temperatures are the same. Thermal equilibrium is established when two bodies are in thermal contact with each other\u2014meaning heat transfer (i.e., the transfer of energy by heat) can occur between them. If two systems cannot freely exchange energy, they will not reach thermal equilibrium. (It is fortunate that empty space stands between Earth and the sun, because a state of thermal equilibrium with the sun would be too toasty for life on this planet!) If two systems, A and B, are in thermal equilibrium with each another, and B is in thermal equilibrium with a third system, C, then A is also in thermal equilibrium with C. This statement may seem obvious, because all three have the same temperature, but it is basic to thermodynamics. It is called the zeroth law of thermodynamics. You may be wondering at this point, why the wacky name? Shouldn\u2019t this be called the first law of thermodynamics rather than the zeroth? The explanation is that this law was discovered after the first and second laws of thermodynamics but is so fundamental that scientists decided it should logically come first. As an example of the zeroth law in action, consider newborn babies in neonatal intensive-care units in hospitals. Prematurely born or sick newborns are placed in special incubators. These babies have very little covering while in the incubators, so to an observer, they look as though they may not be warm enough. However, inside the incubator, the temperature of the air, the cot, and the baby are all the same\u2014that is, they are in thermal equilibrium. The ambient temperature is just high enough to keep the baby safe and comfortable.            
               \end{tabular} &
  \begin{tabular}{@{}p{5cm}@{}}
               \textbf{Chapter Title}:12-1-zeroth-law-of-thermodynamics-thermal-equilibrium \\
\textbf{Learning Objectives}: Explain the zeroth law of thermodynamics \\
\textbf{Summary}:  Systems are in thermal equilibrium when they have the same temperature. Thermal equilibrium occurs when two bodies are in contact with each other and can freely exchange energy. The zeroth law of thermodynamics states that when two systems, A and B, are in thermal equilibrium with each other, and B is in thermal equilibrium with a third system, C, then A is also in thermal equilibrium with C.\\
\textbf{Bold Terms}: thermal equilibrium, zeroth law of thermodynamics \\
\textbf{Key Concepts}: Boltzmann constant, cyclical process, entropy, first law of thermodynamics, heat engine, heat pump, ideal gas law, internal energy, pressure, second law of thermodynamics, thermal efficiency, thermal equilibrium, zeroth law of thermodynamics
               \end{tabular} \\
\end{tabular}
\caption{Example context and formatting information: all dialogues in \cref{tab:dialogue_example_all} are based on the same context and formatting information in this table.}
\label{tab:context_formatting}
\end{table*}

\begin{table*}[h!]
\centering
\small %
\setlength{\tabcolsep}{2pt} %
\begin{tabular}{p{4cm}|p{4cm}|p{8cm}}

SimSeek & Dialogue Inpainting & Persona(High Info) \\
\hline
\begin{tabular}{@{}p{4cm}@{}}
\textbf{S}: What does it mean that they are in thermoequilibrium? \\
\textbf{T}: Heat will transfer thermal energy from the object at higher temperature to the one at lower temperature until they both reach the same temperature. \\
\textbf{S}: What other conditions are required to be in thermoequilibrium? \\
\textbf{T}: The objects are then in thermal equilibrium, and no further temperature changes will occur if they are isolated from other systems. \\
\textbf{S}: What does this mean for the two objects in contact with each other? \\
\textbf{T}: Heat will transfer thermal energy from the object at higher temperature to the one at lower temperature until they both reach the same temperature. \\
\textbf{S}: What else is in thermi-equilibrium? \\
\textbf{T}: The objects are then in thermal equilibrium, and no further temperature changes will occur if they are isolated from other systems. \\
\textbf{S}: What other conditions are required for thermi-equilibrium? \\
\textbf{T}: The systems interact and change because their temperatures are different, and the changes stop once their temperatures are the same. \\
\textbf{S}: What is the other term for the temperature difference in a thermoequilibrium system? \\
\textbf{T}: Heat will transfer thermal energy from the object at higher temperature to the one at lower temperature until they both reach the same temperature.

 \end{tabular}
 &
\begin{tabular}{@{}p{4cm}@{}}
\textbf{S}: Let me see what the relevant information is. \\
\textbf{T}: We learned in the previous chapter that when two objects (or systems) are in contact with one another, heat will transfer thermal energy from the object at higher temperature to the one at lower temperature until they both reach the same temperature. \\
\textbf{S}: What does that mean for the therms in isolation? \\
\textbf{T}: The objects are then in thermal equilibrium, and no further temperature changes will occur if they are isolated from other systems. \\
\textbf{S}: Why are they in thermal equilibrium? \\
\textbf{T}: The systems interact and change because their temperatures are different, and the changes stop once their temperatures are the same. \\
\textbf{S}: How does it work? \\
\textbf{T}: Thermal equilibrium is established when two bodies are in thermal contact with each other\u2014meaning heat transfer (i.e., the transfer of energy by heat) can occur between them. \\
\textbf{S}: Why is that important? \\
\textbf{T}: If two systems cannot freely exchange energy, they will not reach thermal equilibrium. \\
\textbf{S}: What is the most important part? \\
\textbf{T}: (It is fortunate that empty space stands between Earth and the sun, because a state of thermal equilibrium with the sun would be too toasty for life on this planet!) The ambient temperature is just high enough to keep the baby safe and comfortable.

 \end{tabular}
 &
 \begin{tabular}{@{}p{8cm}@{}}
\textbf{S}: What is the zeroth law of thermodynamics? \\
\textbf{T}: The zeroth law of thermodynamics states that if two systems, A and B, are in thermal equilibrium with each other, and B is in thermal equilibrium with a third system, C, then A is also in thermal equilibrium with C. It is called the zeroth law because it was discovered after the first and second laws of thermodynamics, but is considered fundamental and logically comes first. \\
\textbf{S}: Can you explain why the zeroth law of thermodynamics is considered fundamental and logically comes first, even though it was discovered after the first and second laws of thermodynamics? \\
\textbf{T}: The zeroth law of thermodynamics is considered fundamental and logically comes first because it establishes the concept of thermal equilibrium. It states that if two systems are in thermal equilibrium with each other, and one of them is in thermal equilibrium with a third system, then the remaining two systems are also in thermal equilibrium with each other. This law is foundational to thermodynamics because it allows us to define and understand temperature. \\
\textbf{S}: What is the significance of the zeroth law of thermodynamics in defining and understanding temperature? \\
\textbf{T}: The zeroth law of thermodynamics is considered fundamental and logically comes first because it establishes the concept of thermal equilibrium. It states that if two systems are in thermal equilibrium with each other, and one of them is in thermal equilibrium with a third system, then the remaining two systems are also in thermal equilibrium with each other. This law is foundational to thermodynamics because it allows us to define and understand temperature. \\
\textbf{S}: Can you explain the concept of thermal equilibrium and its relationship to the zeroth law of thermodynamics? \\
\textbf{T}: The zeroth law of thermodynamics is considered fundamental and logically comes first because it establishes the concept of thermal equilibrium. This law states that if two systems are in thermal equilibrium with each other, and one of them is in thermal equilibrium with a third system, then the remaining two systems are also in thermal equilibrium with each other. The significance of this law in defining and understanding temperature is that it allows us to determine and compare the temperatures of different systems. It provides a foundation for the measurement and study of heat transfer and the behavior of systems in relation to their temperature. \\
\textbf{S}: How does the zeroth law of thermodynamics allow us to determine and compare the temperatures of different systems? \\
\textbf{T}: The zeroth law of thermodynamics allows us to determine and compare the temperatures of different systems by establishing the concept of thermal equilibrium. Thermal equilibrium occurs when two systems are in contact with each other and have reached the same temperature, resulting in no further temperature changes. If two systems are in thermal equilibrium with each other, and one of them is also in thermal equilibrium with a third system, then the remaining two systems are also in thermal equilibrium with each other. This law is fundamental to thermodynamics as it provides a foundation for measuring and studying heat transfer and the behavior of systems in relation to their temperature. \\
\textbf{S}: How does the zeroth law of thermodynamics define thermal equilibrium? \\
\textbf{T}: The zeroth law of thermodynamics allows us to determine and compare the temperatures of different systems by establishing the concept of thermal equilibrium. Thermal equilibrium occurs when two systems are in contact with each other and have reached the same temperature, resulting in no further temperature changes. If two systems are in thermal equilibrium with each other, and one of them is also in thermal equilibrium with a third system, then the remaining two systems are also in thermal equilibrium with each other. This law is fundamental to thermodynamics as it provides a foundation for measuring and studying heat transfer and the behavior of systems in relation to their temperature.

 \end{tabular}

\end{tabular}
\caption{Comparison of Generated Dialogues: S and T means student and teacher, The Persona model generates more verbose dialogues compared to the other two methods.}
\label{tab:dialogue_example_all}
\end{table*}

\end{document}